\documentclass{article}
\usepackage[utf8]{inputenc}

\usepackage{hyperref}
\usepackage{url}

\usepackage{rotating}      %
\usepackage[utf8]{inputenc} %
\usepackage[T1]{fontenc}    %
\usepackage{hyperref}       %
\usepackage{url}            %
\usepackage{booktabs}       %
\usepackage{amsfonts}       %
\usepackage{nicefrac}       %
\usepackage{microtype}      %

\usepackage{color,xcolor}
\usepackage{enumitem}
\usepackage{bm}
\usepackage{multirow}
\usepackage{bbm}
\usepackage{amsmath} 
\usepackage{algorithm} 
\usepackage[noend]{algpseudocode}
\usepackage{makecell}
\usepackage{tikz}
\usetikzlibrary{spy}

\usepackage[normalem]{ulem}

\newcommand{\sidecap}[1]{ {\begin{sideways}\parbox{1.6cm}{\centering #1}\end{sideways}} }
\newcommand{\sidecapcustom}[2]{ {\begin{sideways}\parbox{#1}{\centering #2}\end{sideways}} }

\usepackage{graphicx}
\graphicspath{{ieee_newimages}}

\def\bmq{\bm{q}}

\def\calI{{\mathcal{I}}}

\def\qu#1{(#1)}
\def\qut#1{\left(#1\right)}
\def\qutb#1{\left[#1\right]}
\def\qutl#1{\mathopen{}\left(#1\right)\mathclose{}}
\def\bmth{\bm{\theta}}

\def\bmsk#1{\bm{s}^{(#1)}}
\def\bmpk#1{\bm{f}^{(#1)}} %

\def\sk#1{{s}^{(#1)}}
\def\hk#1{{h}^{(#1)}}

\def\calI{{\mathcal{I}}}

\usepackage{authblk}

\title{Multi-task deep learning for image segmentation using recursive approximation tasks}
\author[1]{Rihuan Ke}
\author[2]{Aur\'elie Bugeau}
\author[3]{Nicolas Papadakis}
\author[4]{Mark Kirkland}
\author[4]{Peter Schuetz}
\author[1]{Carola-Bibiane Sch{\"o}nlieb}
\affil[1]{DAMTP, University of Cambridge}
\affil[2]{LaBR,   University of  Bordeaux}
\affil[3]{CNRS, IMB}
\affil[4]{Unilever R\&D Colworth}

\date{}

\usepackage[nottoc]{tocbibind}
\usepackage{graphicx}
\usepackage{placeins}

\usepackage[verbose=true,letterpaper]{geometry}
  \newgeometry{
    textheight=9in,
    textwidth=5.5in,
    top=1in,
    headheight=12pt,
    headsep=25pt,
    footskip=30pt
  }

\begin{document}

\maketitle

\begin{abstract}
Fully supervised deep neural networks for segmentation usually require a massive amount of pixel-level labels which are manually expensive to create. In this work, we develop a multi-task learning method to relax this constraint. 
{We regard the segmentation problem as a sequence of approximation subproblems that are recursively defined and in {increasing} levels of approximation accuracy.} The subproblems are handled by a framework that consists of 1) a segmentation task that learns from pixel-level ground truth segmentation masks of a small fraction of the images, 2) a recursive approximation task {that conducts partial object regions learning and data-driven mask evolution starting from partial masks of each object instance,}
and 3) other problem oriented auxiliary tasks that are trained with sparse annotations and promote the learning of dedicated features. Most training images are only labeled by (rough) partial masks, which do not contain exact object boundaries,
rather than by their full segmentation masks. During the training phase, the approximation task learns the statistics of these partial masks, and the partial regions are recursively increased towards object boundaries aided by {the learned information from the} segmentation task {in a fully data-driven fashion}.  The network is trained on an extremely small amount of precisely segmented images and a large set of coarse labels. Annotations can thus be obtained in a cheap way. {We demonstrate the efficiency of our approach in three applications with {microscopy images and ultrasound images.}}
\end{abstract}

\section{Introduction}

The past {years} have seen the success of deep convolutional neural networks (DCNN)  for semantic segmentation \cite{litjens2017survey, long2015fully, chen2018deeplab, badrinarayanan2017segnet}. 
Typical DCNN for segmentation are trained with ground truth segmentation masks in which each pixel of the image is labeled. 
The need for a large set of ground truth segmentation masks, however, hinders the application of DCNN in many application domains where the collection of such masks are  labour intensive and time-consuming. 

There has been a growing interest in the weakly supervised methods for semantic segmentation \cite{huang2018weakly, khoreva2017simple,tsutsui2018minimizing, sun2019saliency}. These methods use weaker forms of annotations such as image-level labels, points-level labels or bounding boxes during the training of DCNN. The weaker annotations (weak labels) are easier to create compared to the ground truth masks (strong labels) and therefore reduce the overall annotation cost. 

\begin{figure}[ht]
  \centering
\includegraphics[width=0.8\linewidth, trim=30  10 30 5 ,clip]{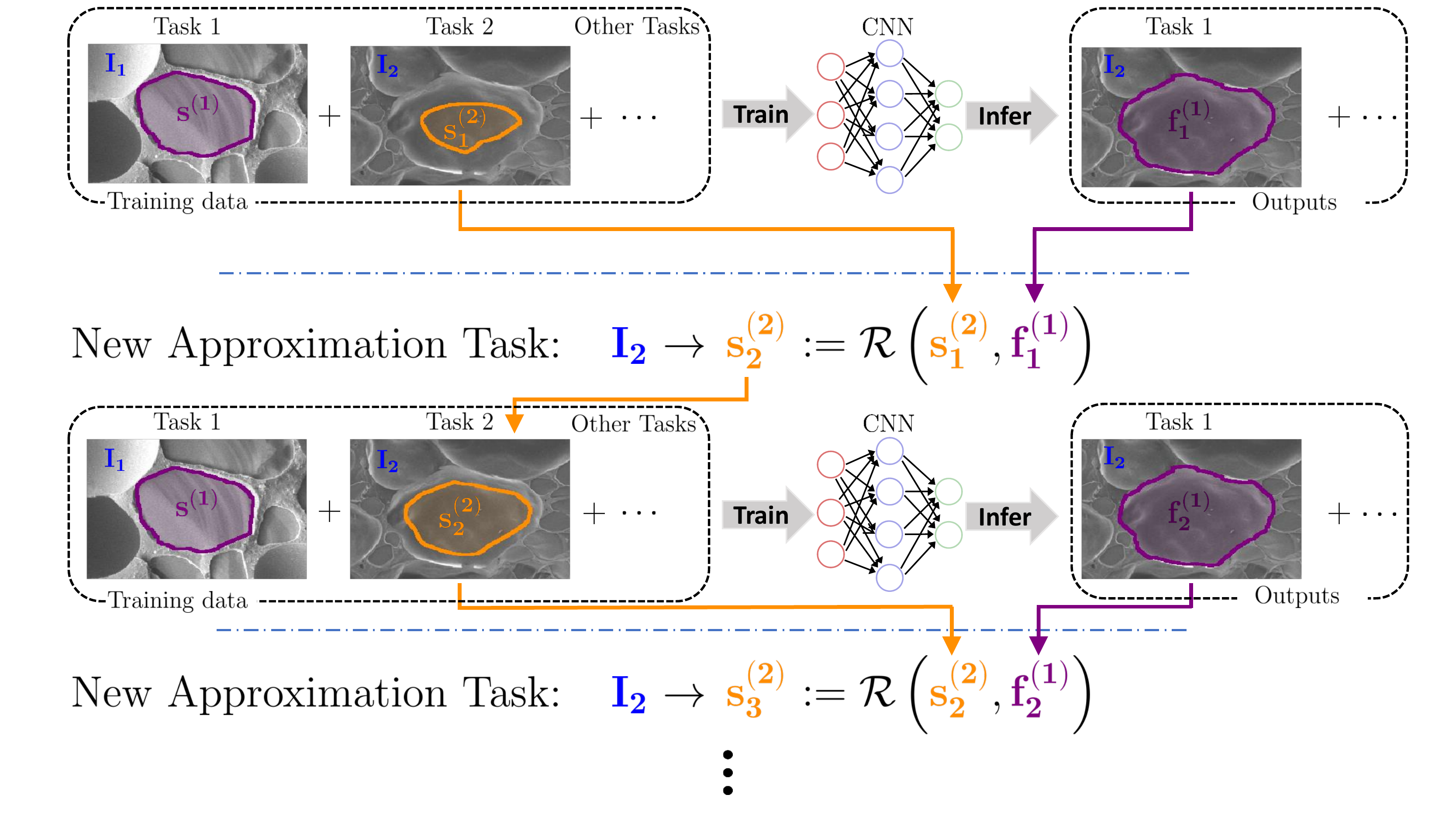}
\vspace{-0.5cm}
  \caption{
  Multi-task learning for image segmentation with recursively updated approximation tasks {(best viewed in color)}. 
  The framework consists of {multiple tasks and} a sequence of iterations on the mask approximation learning {(represented by the rows between the blue dash dot lines).} Each new iteration aims at predicting a larger region of the object to get a mask that better matches object contours (in orange). 
The sets of labeled images do not need to be equal among the tasks. Indeed, only a small fraction of the images are pixel-level annotated (Task 1) and the majority of the images are coarsely annotated (Task 2, Task 3, $\cdots$), our framework needs a much smaller annotation cost compared to the fully supervised approaches.
}\label{fig:icecreamimages}
\end{figure}

{In this paper, {we study the problem of localizing accurate object contours from coarse labels, in a data-driven setting, especially for poorly contrasted images or objects with complex boundaries.} We regard the segmentation problem as a sequence of approximation subproblems and propose a recursive approximation task that learns coarse and cheap labels to approximate the segmentation masks, {initialized by their rough inner regions}. 
{To this end, we introduce several connected tasks (cf. Figure \ref{fig:icecreamimages}). Task 1 is the \textit{segmentation task} trained on \textbf{very few images} with ground truth segmentation masks. Task 2 is the \textit{approximation task} for predicting \textbf{coarse partial masks} given by annotators or their improved versions generated during the learning process.  
Depending on the problem at hand, further \textit{auxiliary tasks} being trained on \textbf{sparse annotations} can be {optionally} incorporated in the framework in order to improve the learning of specific relevant features.}

The approximation task is updated within a multi-task learning loop that integrates all tasks, as demonstrated in Figure \ref{fig:icecreamimages}. 
The tasks are jointly and recursively learned. 
At each learning iteration, the aim of the segmentation task and the additional auxiliary tasks remain unchanged as these tasks are trained with the same data (image and their labels) and try to fit the same probability distributions. The approximation task, in contrast, is updated during the iterations with the targeted regions growing closer to the object boundaries (cf. $\bmsk{2}_k$, $\small k=1,2,\cdots$ in Figure \ref{fig:icecreamimages}).
{We call the masks of these targeted regions \textit{approximations} of the segmentation masks and accordingly refer to the problem of learning such approximations as \textit{approximation tasks}.
}
{The whole learning process} is achieved by connecting the tasks.
The approximation task helps to learn the segmentation, while the learned segmentation task improves the approximations in return.
Taking inspiration from region growing methods in image segmentation, we here propose to incorporate the growth in a learning setting. 
Clearly, in an ideal situation where all the approximations become very accurate, or equivalently, a whole set of ground truth segmentation masks are available for the whole training set, then it reduces to a fully supervised learning setting.} 

{
Hence, the learning framework uses a limited amount of segmentation masks that are available for a small fraction of training images, as well as a large set of cheap and coarse labels such as very rough inner regions given by the annotators (cf. second column of Figure \ref{fig:icecreamimages}). This is motivated by the fact that it is much easier to manually draw a coarse partial mask than delineating a segmentation mask accurately up to a single pixel at the object boundaries. Removing the constraint of fitting object boundaries given by accurate segmentation mask allows freedom of labeling and leave the problems of finding object contours to the underlying network.
}

Compared to existing forms of weak labels for semantic segmentation (e.g., image-level labels, point-level labels, bounding boxes), we use coarse partial masks as approximations to the segmentation masks. The creation of approximations is {flexible}, allowing the annotators to introduce randomness.  {Note that in contrast to the ground truth masks,  initial approximation labels are not uniquely defined given the images.}
The proposed framework uses a learning strategy different from the pseudo label approaches (see e.g., \cite{wei2017stc,khoreva2017simple, lee2019ficklenet, wei2018revisiting, tsutsui2018minimizing}), in which pseudo segmentation masks are generated from the weak labels and are then fed to the segmentation network in a single task manner. {The training therefore can be misled by the wrongly labeled pixels in the pseudo masks or even biased if the pseudo masks are not generated properly.} In this work, however, we {model} different types of labels {separately} {with} different tasks and let the network exploit the statistics of the weak labels and task connections. {The learned features from those statistics are shared {among the} tasks.} {Moreover, instead of having a single-step pseudo mask creation that may foster unreliable object boundaries, we {separate the coarse labels from the segmentation task since the beginning of the learning without a prepossessing step and} improve  {them} gradually with the self-adapted mask approximation and evolution process.} The method is also different from weakly supervised methods using a composite loss (see e.g., \cite{kolesnikov2016seed, huang2018weakly, sun2019saliency}) that forces the network to predict a segmentation mask that respects all weak labels. These methods may ignore some important information {well} contained in weak labels {such as the boundaries of two connected object instances.}

The highlights of this paper are summarized as follows. 
(1) We propose a new multi-task learning framework that splits into {a list of less annotation-demanding tasks.} The weak labels are used for feeding the network and they require no preprocessing effort. 
(2) We introduce {the recursive approximation task that conducts sequential mask approximations at increasing accuracy levels.}
{The framework integrates region growing in a completely data-driven manner.}
(3) The efficiency of the proposed learning strategy is demonstrated in {three applications in segmenting microscopy images and ultrasound images}. 

\section{Related Work}%
Image segmentation is an important subject in computer vision with many real world applications. There exist a diversity of algorithms for segmentation, ranging from unsupervised methods like thresholding and edge detection to the recent deep learning approaches. The traditional intensity or region based segmentation approaches are usually very computationally efficient and they require no resources in the form of manual annotations. Nevertheless,  they are restricted to the underlying low level image attributes such as pixel intensity and fail to handle images in complex scenarios.  Deep learning segmentation methods have gained great popularity in recent years for the semantic segmentation problems. 
The proposed method is built on top of the works of both categories that are now reviewed.
\subsection{Region growing and contour evolution}
Seeded region growing methods \cite{adams1994seeded, zhu1996region} are one of the most classical image segmentation approaches. Those methods are initialized with labeled regions called seeds, either obtained manually or in an automated manner. Unlabeled neighboring pixels are iteratively  merged with the seeds. The decision on which pixels should be merged to the seed is made upon a criterion that measures the distance between a pixel and the seeded regions. One typical criterion is the absolute value of the distance between the intensity of neighboring pixels (see e.g., \cite{adams1994seeded, mehnert1997improved}).
Region growing methods are known to be powerful in grouping homogeneous areas of the images. However, if the similarity criterion is violated, for instance {in the presence of textures, artifacts or low contrast between objects and backgrounds,} they may result in segmentations of poor quality. 

Active contour methods \cite{kass1988snakes,caselles1993geometric, caselles1997geodesic} belong to another family of techniques that iteratively update a curve to match object boundaries by minimizing a given energy function.
The earliest active contour method  \cite{kass1988snakes}  gathers an internal energy term that imposes the regularity on the parameterized curve as well as an image force that constrains the curve to fit object boundaries. 
Active contours models can also be formulated with level sets \cite{osher1988fronts}, where the curve evolution is modeled as the propagation of the zero level set of the signed distance function through a velocity field defined by local color features \cite{caselles1993geometric,kichenassamy1996conformal,caselles1997geodesic}.

\subsection{Weakly supervised learning for segmentation}
The recent development of the deep learning approaches has shown their capacity to learn features for complex segmentation tasks. 
Instead of relying on predefined low-level features, recent deep learning approaches have proposed to learn features for complex segmentation tasks.
Different forms of supervision are developed for deep segmentation networks. Fully supervised learning, which relies on pixel-level ground truth annotations, is among the most standard and widely investigated learning scheme for segmentation {(see e.g., \cite{badrinarayanan2017segnet, ronneberger2015u, chen2018deeplab})}. Weakly supervised learning methods  use weaker forms of annotations for segmentation, that can be obtained more efficiently. Different kinds of weak supervisions, depending on the available type of annotations, also exist in the literature.  Typical weak annotations are points \cite{bearman2016s}, scribbles \cite{lin2016scribblesup}, bounding boxes \cite{khoreva2017simple} and image-level tags \cite{huang2018weakly, papandreou2015weakly,lee2019ficklenet}.

Image classification and segmentation are two closely connected problems. Image-level labels are probably the simplest type of weak annotations for segmentation learning. Such annotations can be used for training classification networks, from which the learned features are found to be useful for learning how to segment images (see e.g., \cite{long2015fully}). An Expectation-Maximization approach  is proposed in \cite{papandreou2015weakly} to train a segmentation model using image-level labels. {The approach in \cite{kolesnikov2016seed} introduces} a composite loss function that forces the network to predict segmentation masks corresponding to image level labels, seed cues generated from the classification networks, and a boundary constraint. Learning with image level labels can be improved with the help of unsupervised approaches. The STC (simple to complex) framework \cite{wei2017stc} employs the saliency maps obtained from unsupervised saliency detection methods to train a segmentation network which is then adjusted with the image level labels. 
Since the cues from image level supervision are usually not accurate enough at object boundaries, post-processing techniques, such as conditional random fields, are often applied to improve segmentation results. 
Several authors exploit seeded region growing to expand the seed cues for better supervision \cite{huang2018weakly, sun2019saliency}. 

Pseudo label (PL) based techniques have been designed for training segmentation networks from weak annotations. 
PL methods require to generate  pseudo segmentation masks that are considered as ground truth during the training. In \cite{khoreva2017simple}, the PL masks are computed with Grabcut \cite{rother2004grabcut} from bounding boxes.
The learning strategy in \cite{tsutsui2018minimizing} obtains the PL with $k$-means and and a location prior in the images. 

In contrast to PL based approaches, our method learns an approximation task starting from the weak labels without a preprocessing step. Better approximations are provided by the network itself during the training. To do so, we jointly learn to segment images and to improve the accuracy of the rough annotations given in the training data. The two objectives are treated through a recursive approximation task that promotes the consistency of both predicted results. 

{Our work is relevant to the multi-task method in \cite{playout2019novel}. It integrates a detection task in a weakly supervised manner to help the segmentation of retinal lesions. 
A related idea has been investigated in \cite{mlynarski2019deep} where a segmentation task is jointly learned with a classification task, supervised by image-level labels, for brain tumor segmentation. The work \cite{perone2018deep}, in contrast, explores a consistency loss to leverage unlabeled data for the segmentation of magnetic resonance imaging data via a mean teacher approach.}

\section{The proposed multi-task learning method}\label{sec:model}

{
Multi-task learning takes advantage of the information shared among two or more connected tasks for better handling the tasks. 
The framework developed in this work consists of a segmentation task, an approximation task and additional auxiliary tasks (cf. Figure \ref{fig:icecreamimages}). For each task, associated labels are required in the learning process. In this section we will first formulate the tasks and the associated labels, and then introduce the learning method for the recursive approximation task.}

{Let {\small $\calI$} be a set of images and consider $T$ different tasks ($T \geq 2$).} 
For each task, paired examples are given in the form 
\[
\qut{I, \bmsk{t}_k},  \quad I \in \calI_t, \quad t \in \{1, 2, \cdots, T\},
\]
where $\calI_t \subset{\calI}$ is the subset of images labeled for Task $t$, and $\bmsk{t}_k$ is the targeted output of Task $t$ given the image $I$. 
We consider here a dynamic learning process where the targets depend on the iteration number denoted by $k$.
Supervised learning methods are designed to predict $\bmsk{t}_k$ from a new image $I$.

The main challenge tackled in this work is to perform an accurate segmentation and match object contours, given that only a very small amount of images are provided with ground truth segmentation masks. We decompose the segmentation problem in the following way. 
First, \textit{Task 1} is the segmentation task and thus $\bmsk{1}_k$ denotes the segmentation mask (cf., first column in Figure \ref{fig:icecreamimages}). The mask $\bmsk{1}_k$ is available only for a small fraction of the whole set of images, i.e., $|\calI_1| \ll | \calI |$. For the segmentation task we assume $C_1$ classes and a one-hot representation of the mask, i.e., $[\bmsk{1}_k]_{i,c}=1$ if the pixel $i$ of image $I$ belongs to the class $c$, and otherwise $[\bmsk{1}_k]_{i,c}=0$ {for $0 \leq c \leq C_1 - 1$}. 
Second, \textit{Task 2} is the approximation task, in which $\bmsk{2}_k$ represents the partial mask for the object instances, cf. the second column of Figure \ref{fig:icecreamimages}. Usually the initial label $\bmsk{2}_1$ describes the inner regions of the object mask $\bmsk{1}_k$ but does not capture the object contours, i.e.  they are missing a lot of pixels at the boundaries of the objects.
The approximation $\bmsk{2}_k$ has the same spatial dimensionality as the image $I$ and the mask $\bmsk{1}_k$.
Finally, \textit{Task 3, $\cdots$, T} are other auxiliary tasks with coarse labels denoted as $\bmsk{t}_k$. 
{
All tasks (including the segmentation and approximation tasks) discussed in this work are pixel-wise classification problems in which the targets are $\bmsk{t}_k$ containing $C_t$ classes. 
{Therefore, analogously to Task 1, for any $t\geq 2$} the entries of the labels for task $t$ are denoted by $[\bmsk{t}_k]_{i,c}$ where $i$ is the index for image pixels and $c = 0, 1, \cdots, C_{t}-1$.}

If we assume that the image $I$ is randomly drawn from a distribution on $\calI$, then we are interested in knowing the conditional probability distribution of the mask $\bmsk{1}_k$ given $I$, described by $p\qu{\bmsk{1}_k \mid  I }$. Fully supervised methods learn the distribution with {labelled pairs} $\qu{I, \bmsk{t}_k}$ for all $I$ across the whole training set. Finding {$p\qu{\bmsk{1}_k \mid  I }$} becomes even more challenging if the amount of such pairs is limited to a small subset of the whole dataset.

The approximation task, as its name suggests, approximates the mask $\bmsk{1}_k$ and helps to learn {$p\qu{\bmsk{1}_k \mid  I }$}. {The approximate labels also bring a possibility of training an approximate model for segmentation, which predicts accurately at least in the inner regions of objects. Such an approximate model provides a starting point for segmentation or for getting better approximations. For instance, in the setting of DCNN the learned features for this task can be reused for the segmentation task.} {Note that these inaccurate labels $\bmsk{2}_k$, however, are unseen by Task $1$ as they are modeled as approximations and treated in Task 2 (i.e., the approximation task).}
We can also define problem oriented tasks $\bmsk{3}_k, \cdots, \bmsk{T}_k$ that highlight specific parts of contents in the images. For example, for images with densely distributed object instances, separating the connected instances can be a difficult problem. Therefore one may define Task 3 as locating the interface between touching objects without clear boundaries between them. It implies a two-classes classification problem and helps to determine a part of $\bmsk{1}_k$.

Provided that parts of the information of $\bmsk{1}_k$ are encoded in $\bmsk{2}_k, \bmsk{3}_k, \cdots, \bmsk{T}_k$, a natural idea is to learn the joint conditional probability 
$
p\qutl{\bmsk{1}_k, \cdots, \bmsk{T}_k \mid  I }
$
instead of individual posterior distributions. As relatively more samples of $\bmsk{2}_k, \bmsk{3}_k, \cdots, \bmsk{T}_k$ are available, learning their statistics is easier. If we assume that $\bmsk{1}_k, \cdots, \bmsk{T}_k$ are conditionally independent given image $I$, then
\[
p\qutl{\bmsk{1}_k, \cdots, \bmsk{T}_k \mid  I } = \prod_{t=1}^T p\qutl{\bmsk{t}_k \mid  I }. 
\]
{Though the tasks are strongly connected as discussed earlier,} this assumption holds in many application settings since the labels for task $t$ can be created independently based on the image $I$ only, without knowing the labels for the other {\small $T-1$} tasks. 

To approximate the joint distribution, consider the model  $p\qu{\bmsk{1}_k, \cdots, \bmsk{T}_k \mid  I; \bmth }$ with parameter $\bmth$ shared among the tasks and the log-likelihood being
\[
\small
\sum_{t=1}^T  {\log p\qutl{\bmsk{t}_k \mid  I; \bmth }}
\]
for the data $(\bmsk{1}_k, \bmsk{2}_k, \cdots, \bmsk{T}_k, I)$. 
Recall that we have an unbalanced set of samples for different tasks, i.e., the labels $\bmsk{t}_k$ are not available for $I$ in $\calI \backslash \calI_t$. We approximate the maximum likelihood (ML) estimate with 
\begin{equation}\label{eq:MLE-2}
\small
\hat{\bmth} = \arg\max_{\bmth}  
\sum_{t=1}^{T}  \qut{ \alpha_t \sum_{I \in \calI_t} {   \log  p\qutl{ \bmsk{t}_k \mid I; \bmth} } }
\end{equation}
in which $\alpha_1, \alpha_2, \cdots, \alpha_T$ are non negative constants.

\subsection{Negative log-likelihood loss}
The optimization problem \eqref{eq:MLE-2} has to be solved to find the parameter $\hat{\bmth}$. 
We use a multi-task DCNN to predict directly $\bmsk{t}_k$ for Task $t$, therefore having $T$ outputs. 
The outputs of the model are respectively represented with the arrays {\small $\hk{1}_{\bm{\theta}}(I)$, $\hk{2}_{\bm{\theta}}(I)$}, $\cdots$, {\small $\hk{T}_{\bm{\theta}}(I)$}. Each array $\hk{t}_{\bm{\theta}}(I)$ has the same size as $s^{(t)}$.  In this work we use the log-likelihood
\begin{equation}\label{logP}
\small
\log  p\qutl{ \bmsk{t}_k \mid I; \bmth}
= \sum_{i} \sum_{0 \leq c \leq C_t - 1}  [\sk{t}_k]_{i,c} \log \qutb{{\hk{t}_{\bm{\theta}}(I)}}_{i,c},
\end{equation}
in which $\qutb{{\hk{t}_{\bm{\theta}}(I)}}_{i,c}$ is the output of the model at pixel location $i$ for the class $c$ of Task $t$. 

Including the log-likelihood  \eqref{logP} into relation (\ref{eq:MLE-2}), the following cross entropy loss is minimized
\begin{equation}\label{eq:loss}
\small
L\qut{\bmth} = - \sum_{I \in \calI} \sum_{t=1}^{T} \alpha_t  \mathbbm{1}_{\calI_t}
\qutl{I}\log  p\qutl{ \bmsk{t}_k \mid I; \bmth},
\end{equation}
in which {\small $ \mathbbm{1}_{\calI_t}\qutl{I} = 1 $} if {\small $I \in \calI_t$} and {\small $0$} if {\small $I \not\in \calI_t$}. 

\subsection{Recursive approximation tasks}\label{subs:recursive}
In Task 2, very coarse masks are used as approximations of the segmentation masks. They describe inner parts of the object regions and thus miss all boundary pixels.
{Region growing  (see e.g., \cite{adams1994seeded, zhu1996region}) approaches are relevant tools to expand such rough initialisation.} The main idea is to let the initial region evolves outwards, with respect to some similarity criterion, and stop when the object contour is reached. 
Local image attributes, such as pixel intensities or gradients, are generally considered. The accuracy of these methods nevertheless strongly relies on the choice of the similarity criterion, which might differ from case to case.  

In our multi-task learning framework,   initial regions are defined from partial masks and they are allowed to grow towards the object boundaries,  to obtain more accurate contours of the objects.  This is done in a data-driven way, by jointly learning the contour information from a small portion of the set of training images {as an additional task}. More specifically, the method learns to predict the segmentation as well as {the initial regions assigned by annotators}. The segmentation results are then used as cues for  region growing. The whole process (cf. Figure \ref{fig:icecreamimages}) consisting of iterations on the approximation tasks and calculation of the new approximation labels via a recursive formula is now introduced. 
\begin{figure}[ht!]
  \centering
\setlength{\tabcolsep}{1pt}
\begin{tabular}{ccc}
\includegraphics[width=0.2\linewidth, trim=30  30 30 30 ,clip]{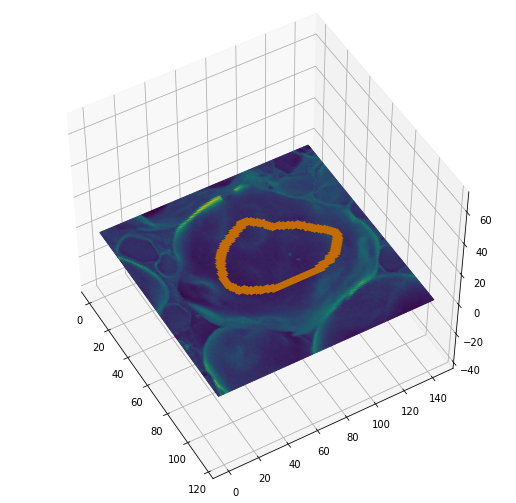}& 
\includegraphics[width=0.2\linewidth, trim=30  30 30 30 ,clip]{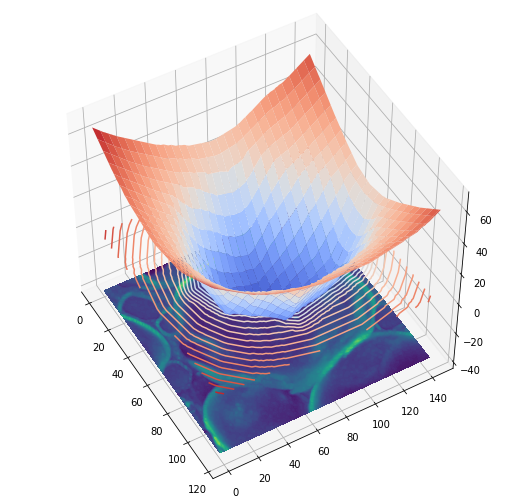}& 
\includegraphics[width=0.2\linewidth, trim=30  30 30 30 ,clip]{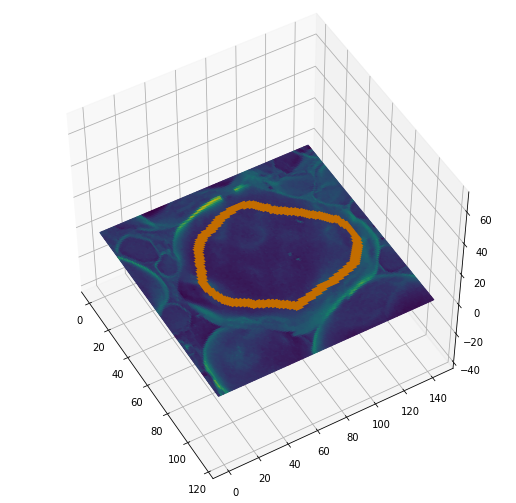}
\\
\multicolumn{3}{c}{\tiny (a) SEM image}\\
\addlinespace[1ex]
\includegraphics[width=0.2\linewidth, trim=30  30 30 30 ,clip]{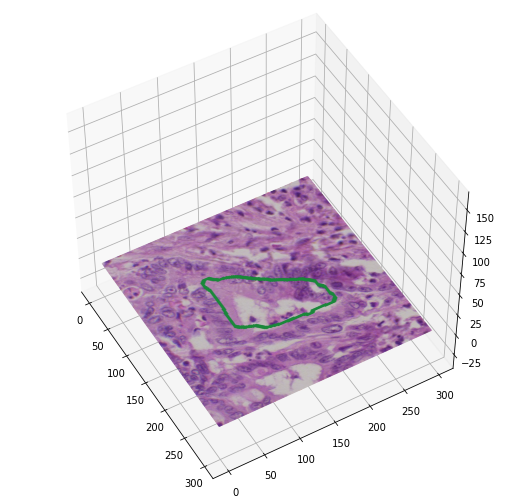}&
\includegraphics[width=0.2\linewidth, trim=30  30 30 30 ,clip]{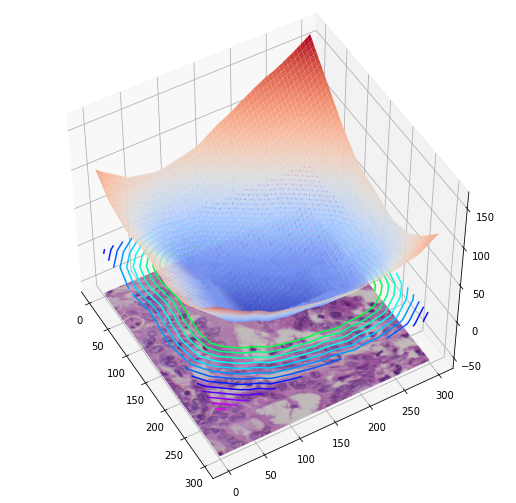}& 
\includegraphics[width=0.2\linewidth, trim=30  30 30 30 ,clip]{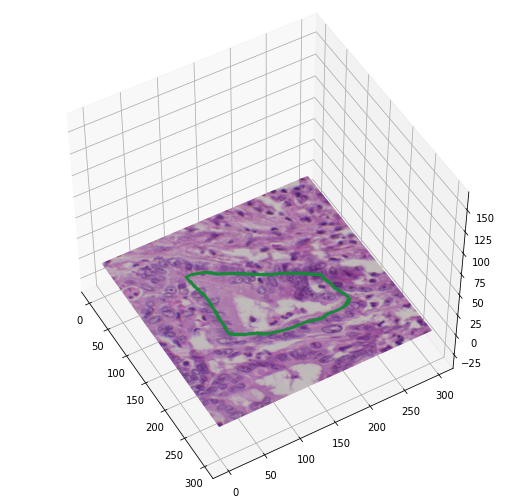}\\
\multicolumn{3}{c}{\tiny (b) H\&E-stained image}
\end{tabular}
  \caption{One step of the recursive approximations demonstrated for two types of images. Left column: the contour (in orange color on the top and green color on the bottom) at iteration $k$. Middle column: the level set function and its contour lines which are computed without handcrafted features. Right column: the new contour used to learn the approximation at iteration $k+1$.}\label{fig:level-set-function}
\end{figure}

We let the labels $\bmsk{t}_k := \bmsk{t}$ for $\small{t \in \{ 1,  \cdots, T\}} \backslash \{2\}$ be fixed with respect to the iteration number $k$, and the associated tasks therefore remain unchanged during the learning phase. The approximation task $2$, in contrast, is updated recursively. During each iteration, the aim is to learn a better approximation, with a larger inner region, based on the results of previous iterations. Let the iteration number be $k$ and $\bmsk{2}_{k-1}$ denote the labels for Task 2 at iteration {\small $k-1$} (cf. orange contours in Figure \ref{fig:icecreamimages}). 
Recall that Task $1$ focuses on object classification and accurate segmentation of their contours{ which are learned together with the approximations.} We also have predictions of the segmentation mask for Task $1$ (c.f., purple regions on the rightmost column of Figure \ref{fig:icecreamimages}) at iteration $\small {k-1}$. They are denoted by $\bmpk{1}_{k-1}$  and the entry $[\bmpk{1}_{k-1}]_{i,c} \in \{0,1\}$ with value $1$ indicating that pixel $i$ belongs to class $c$.
The labeled regions for Task $2$ at iteration $k$ are computed as
\begin{equation}\label{eq:R}
\bmsk{2}_k := \mathcal{R}\qutl{ \bmsk{2}_{k-1}, \bmpk{1}_{k-1}}.
\end{equation}
{The new mask $\bmsk{2}_k$ defines new approximations of the regions of interest that will be learned during next iteration.}

For the sake of clarity, we assume that the number of classes (excluding backgrounds) for Task $1$ and Task $2$ is $1$ and only a single object is contained in the image $I$, then for $[\bmsk{2}_{k}]_{i,c}$ the subscript {\small $c=0$} denotes the object and {\small $c=1$} denotes the background. 
The following recursive formula for approximation can be extended to the multi-classes and multi-objects case.  
We first define a level set function which reads at pixel $i$
\begin{equation}\label{eq:levelset}
\phi\qutl{i} =  {\rm dist}\qutl{i, \sigma\qutl{\bmsk{2}_{k-1}}} - \beta {\rm dist}\qutl{i, \sigma\qutl{1-\bmpk{1}_{k-1}}}
\end{equation}
in which $\sigma\qutl{\bmq} = \left\{ j \mid [\bmq]_{j,0} =1 \right\}$ for the binary mask $\bmq$, $\beta \geq 0$ is a parameter that controls the growth speed, and
\[
\small
{\rm dist}\qutl{i, \sigma\qutl{\bmq}} := 
\min_{j \in \sigma\qutl{\bmq}} {\rm distance}\qutl{i, j}
\]
denotes the distance between $i$ and the set $\sigma\qutl{\bmq}$. 
Then $\mathcal{R}$ is determined by 
\begin{equation}\label{eq:recursive}
\left[\mathcal{R}\qutl{ \bmsk{2}_{k-1}, \bmpk{1}_{k-1}}\right]_{i,0}  = 
\begin{cases}
1, & \text{ if } \phi\qutl{i} \leq 0 \\
0, & \text{otherwise}. 
\end{cases}
\end{equation}

{\noindent The iterative process is dependent on the parameter $\beta$. Since in equation \eqref{eq:levelset} the term ${\rm dist}\qutl{i, \sigma\qutl{1-\bmpk{1}_{k-1}}}$ is non-negative, a larger $\beta$  corresponds to larger object regions in the new approximation masks. If $\beta=0$, then $\bmsk{2}_k$ and $\bmsk{2}_{k-1}$ are identical.  In the other extreme case $\beta = +\infty$, $\mathcal{R}\qutl{ \bmsk{2}_{k-1}, \bmpk{1}_{k-1}}$ is identical to $\bmpk{1}_{k-1}$ under the condition that $[\bmsk{2}_{k-1}]_{i,0} \leq [\bmpk{1}_{k-1}]_{i,0}$ for any $i$. However, recall that $\bmpk{1}_{k-1}$ is an output of the network for Task 1, {and feeding it to Task 2 encourages Task 2 to predict the same as Task $1$, which may be inaccurate at early stages and not very helpful.} In practice, we use small values of $\beta$ such as $1$.}
An example of a computed level set function $\phi\qut{\cdot}$ and the contour specified by $\mathcal{R}\qutl{ \bmsk{2}_{k-1}, \bmpk{1}_{k-1}}$ are respectively shown on the middle and right columns of Figure \ref{fig:level-set-function}.

The recursive approximation learning is summarized in Algorithm \ref{algo:recu}. 

\begin{figure}[ht!]
  \centering
\setlength{\tabcolsep}{1pt}
\begin{tabular}{ccc}
\includegraphics[width=0.3\linewidth, trim=270 140 280 140 ,clip]{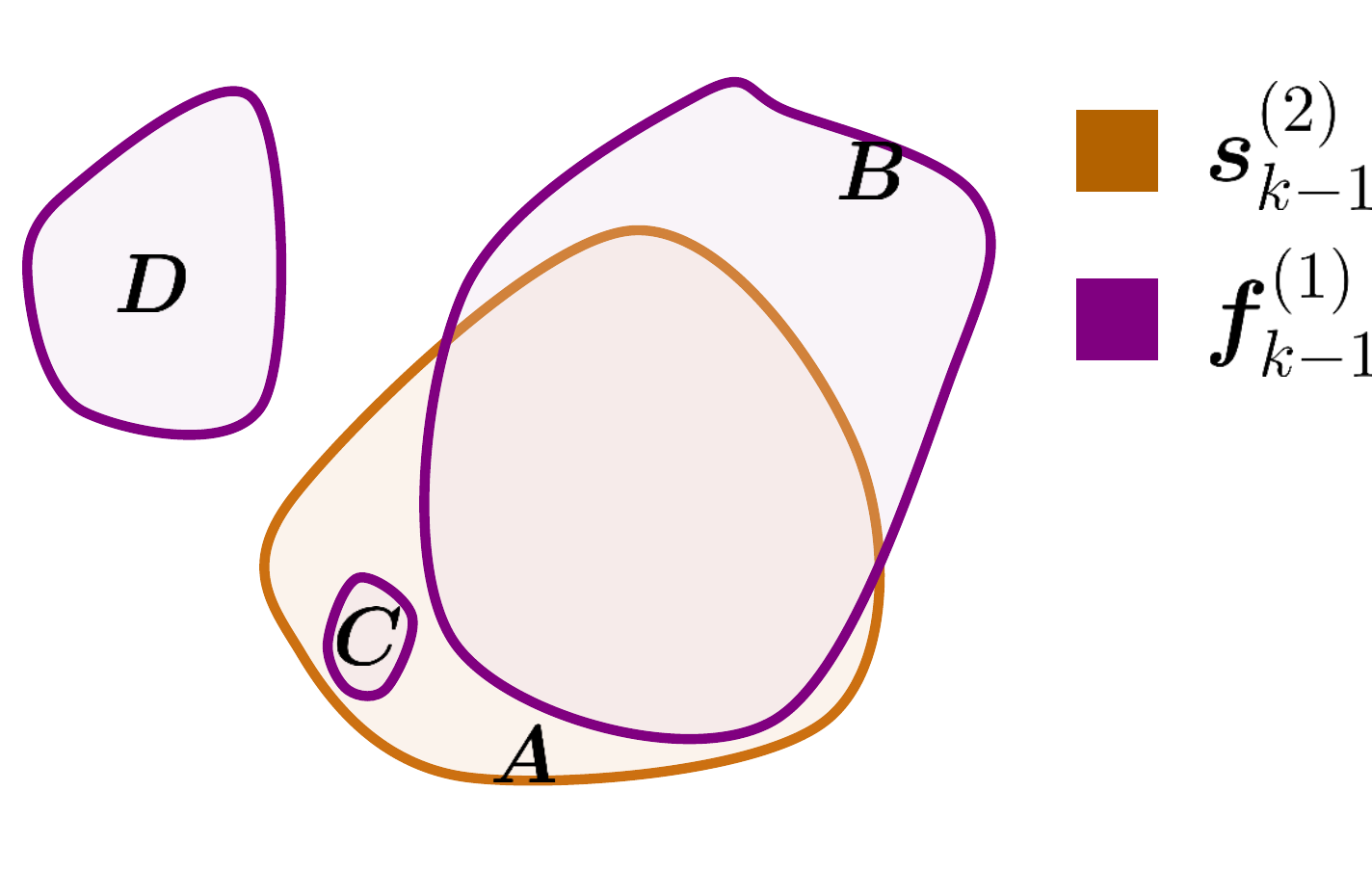}& 
\includegraphics[width=0.3\linewidth, trim=270 140 280 140 ,clip]{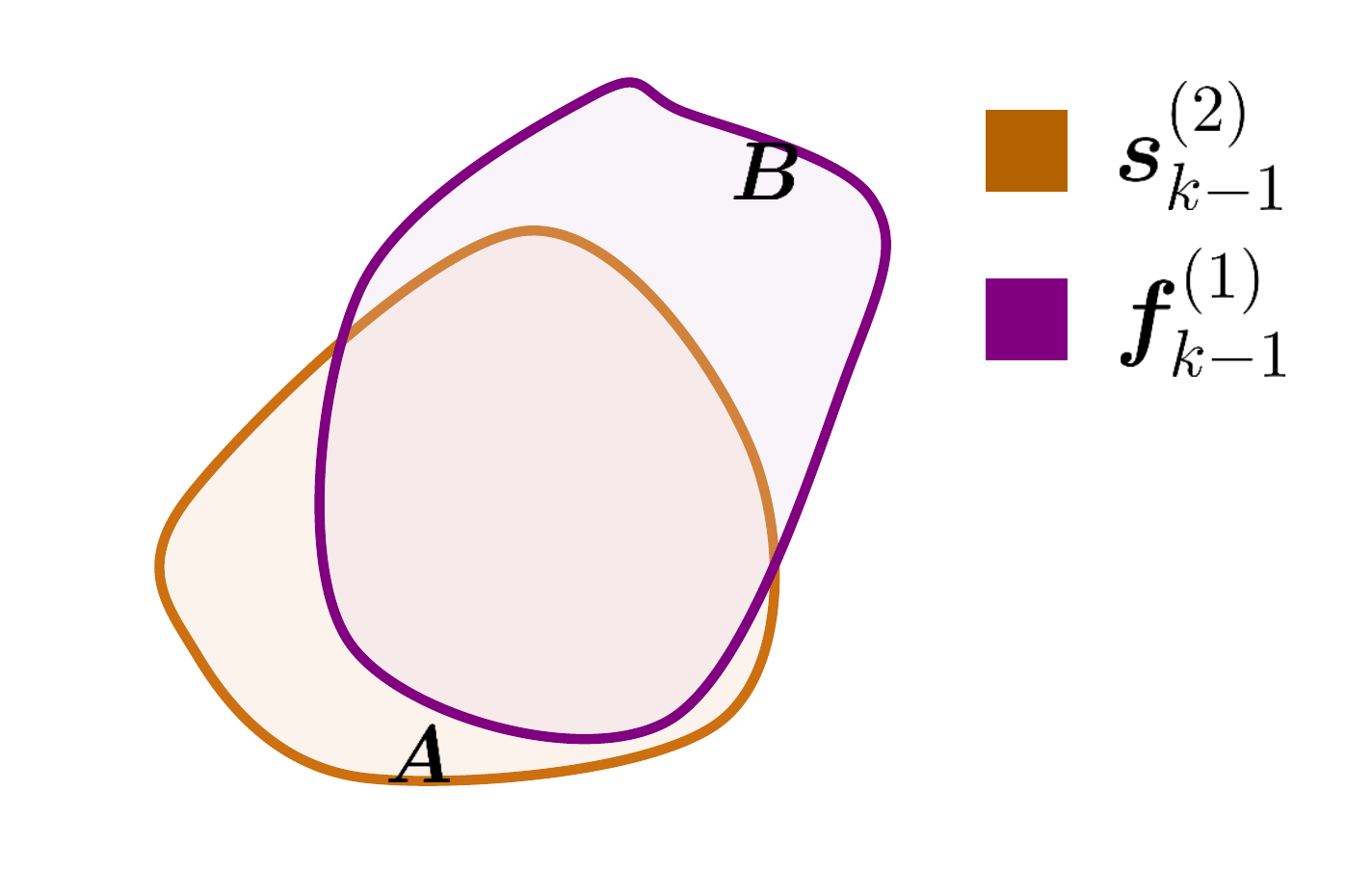} \\
{\footnotesize (a). original regions} & (b). {\footnotesize paired regions}
\end{tabular}
  \caption{Pairing connected components (purple) to seed regions (orange)  for multi-object label updates}\label{fig:region-grow-pairing}
\end{figure}

\begin{algorithm}
\small
\caption{Recursive approximation in multi-task learning}\label{algo:recu}
\begin{algorithmic}[1]
\Procedure{RecursiveApproximation}{$\calI$,  $\bmsk{1},\cdots, \bmsk{T}$, $N$} %
   \State Let  $\bmsk{2}_1 =  \bmsk{2}$ and $k=1$.
   \While{Iteration Number $k \leq N$}
 	\State Minimize the loss function \eqref{eq:loss} with respect to $\bmth$.  
    \State With the new parameter $\bmth$, obtain the predictions of the network $\bmpk{1}_k$ on $\calI$. \\
    \State For each $I \in \calI$ in Task $2$, calculate the new object regions by \newline
    \quad\quad\quad\quad
    \[
    \small{
    \bmsk{2}_{k+1} := \mathcal{R}\qutl{ \bmsk{2}_{k}, \bmpk{1}_{k}}
    }
    \]
    \State $k \leftarrow k + 1$
    \EndWhile
    \State \Return $\bmth$.    
\EndProcedure
\end{algorithmic}
\end{algorithm}
Label update in the multi-object case follows a similar procedure:  a level set function is defined for each pair of masks associated with an individual object.  
This requires an iteration over selected pairs of masks. %
The following pipeline is considered for mulit-object label update.
\begin{enumerate}[label=Step \arabic*.,leftmargin=4\parindent]
\item The seed regions (i.e. partial object regions) represented in $\bmsk{2}_{k-1}$ are extracted. 
\item The connected components (i.e. predicted full object regions) in the mask $\bmpk{1}_{k-1}$ are computed.
\item If a connected component from Step $2$ is overlapping with more than one seed region, then it is split into smaller ones such that each of them intersects with at most one seed region. This can be easily done by removing those pixels lying at the middle of two neighboring regions {from the connected component}. 
\item For a given seed region, if at least $50\%$ of its pixels are covered by a connected component, then it is paired with this connected component (cf., Figure \ref{fig:region-grow-pairing}) and updated by the formula \eqref{eq:R} with $\bmsk{2}_{k-1}$ and  $\bmpk{1}_{k-1}$  being replaced by the mask of the seed region and the mask of the connected component respectively.
\end{enumerate}
The new label $\bmsk{2}_k$ is obtained after all possible pairs are processed. Step 3 guarantees that no objects are merged after region growing. 

Finally, there are various ways to extend the (approximated) label updates into the multi-class and multi-object setting. 
One of the simplest methods is to treat all object classes as a single class (i.e., foreground) and therefore the steps described above can be carried out. The grown regions are then classified using the original class of the seed regions.

\subsection{Preparation of annotations}
Standard fully supervised learning framework requires accurate pixel-level annotations for each image of the training set. In this work, the annotations are created in a different manner. Rough labels (with some randomness) can be provided by the annotators with minimal adjustments or corrections , i.e. without a pseudo label generation step.  The label generation procedure is now illustrated.  

The first step provides  rough approximations of  ground truth segmentation masks, leaving aside the most time-consuming and tedious work, e.g.,  determining labels for the region accurate up to pixels. The annotator may assign to each object instance an initial region that does not necessarily capture the boundaries, cf. middle column of Figure \ref{fig:annotationdemos}.  This can be completely done by manual drawing, given that the contours of the initial regions are not uniquely defined, and we use them as initial approximations of the masks of the objects. In the second step we need to collect ground truth segmentation mask for a relatively small amount of images. One can either start from scratch using any existing methods {to generate the} segmentation mask, or exploit the approximations obtained in the first step. Interactive segmentation techniques can integrate the available initial regions with further manual annotations. Usually iterations of human inputs/corrections are necessary to get the accurate segmentation mask (cf. left column of Figure \ref{fig:annotationdemos}), and the cost per image is much higher than the ones in the first step. Other coarse annotations are potentially generated for Tasks $3$, $\cdots$, Task {\small $T$} if {\small $T \geq 2$}. An example of annotation of a third task is shown in the right column of Figure \ref{fig:annotationdemos}: the connection of neighboring objects without clear separating boundaries are picked out with rough scribbles. Those targeted annotations might be very sparse for some applications, but since the cost per instance is small, a large set of images can be annotated under limited annotation budgets.

\subsection{Network architecture} \label{subs:network}
\begin{figure}
  \centering
\begin{tabular}{c}
    \includegraphics[width=0.6\linewidth, trim=350 140 350 120 ,clip]{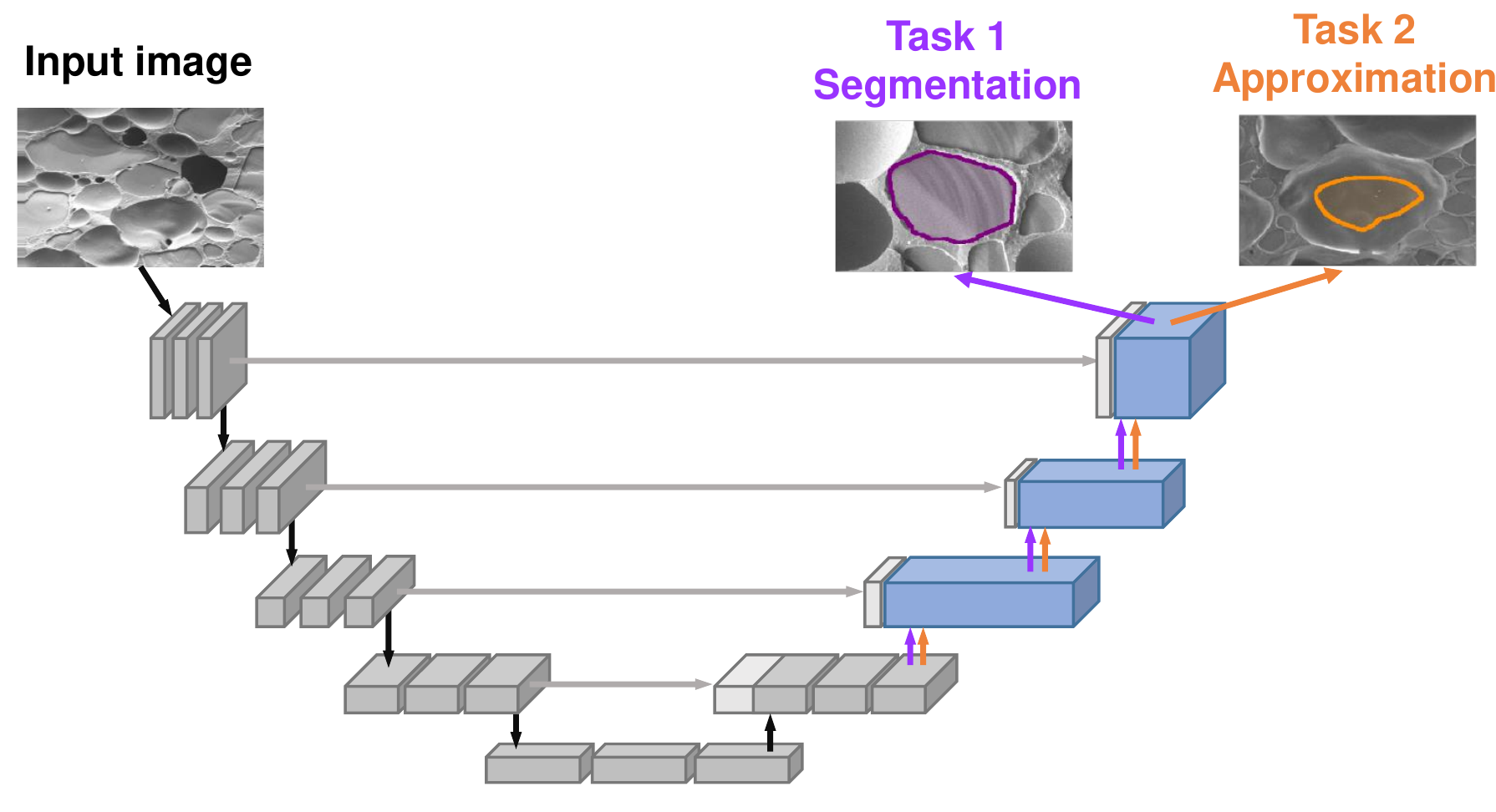} \\
    (a) \\
    \includegraphics[width=0.6\linewidth, trim=330 200 330 180 ,clip]{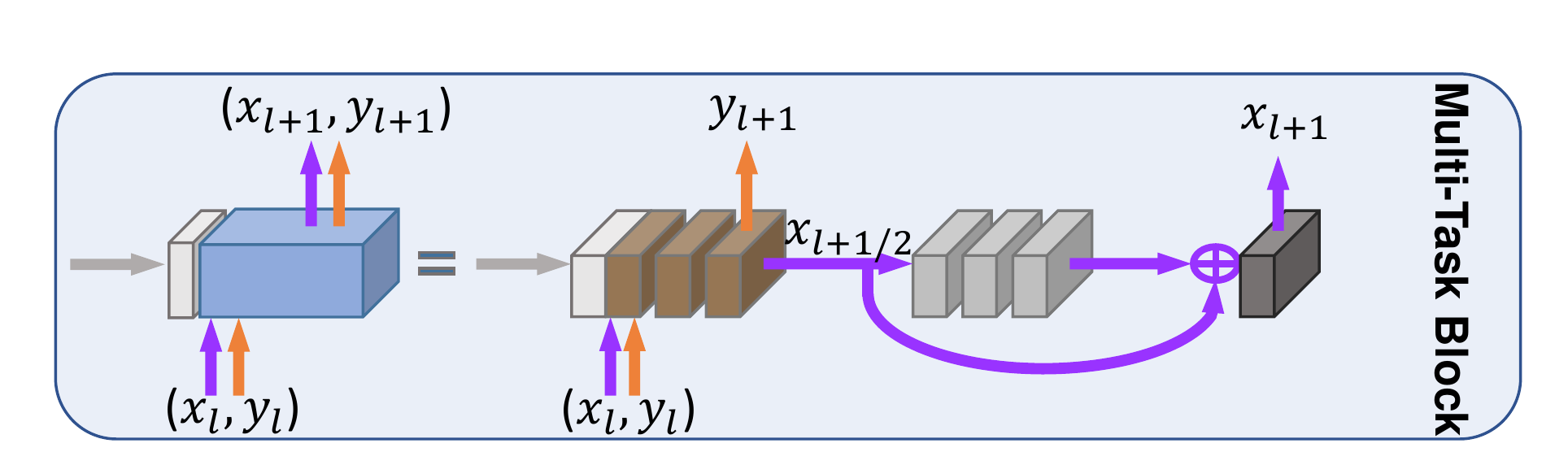}\\
    (b)
\end{tabular}
  \caption{Architecture of the multi-task network (two tasks are shown), inspired by the U-net. (a). 
  The left part (contracting path) of the network is shared by the tasks, 
  and the multi-task blocks on the right part are for learning task specific features. 
  The horizontal gray arrows stand for the long skip connections from the contracting path to the expansive paths via concatenation. The purple lines and orange lines represent respectively the paths for segmentation task and approximation task{, and $x_{l+1}$ (resp. $y_{l+1}$) denotes the output of $l^{\rm th}$ multi-task block for the segmentation task (resp. approximation task)}.
  (b). 
  The two paths run through the multi-task blocks. In the multi-task blocks the first layers of both paths share the same weights (shown in brown). It is followed by a residual block for the segmentation path. 
}\label{fig:network-arch}
\end{figure}

The design of the network architectures has certain flexibility depending on the problem at hand. 
We use a multi-task network adapted from the
well-known U-net \cite{ronneberger2015u} for segmentation. 
The network has an encoder-decoder architecture as shown in Figure \ref{fig:network-arch}. 
Similar to the U-net, it has a contracting path and an expansive path with long-skip connections between them. The resolution of the features on the contracting path is down-sampled 
several times and then up-sampled
on the expansive path successively to the same resolution as the input image. {Down-sampling and up-sampling are indicated respectively by the down arrows and up arrows between blocks in Figure \ref{fig:network-arch}.}
The feature maps on the contracting path are concatenated to the ones on the expansive path to promote the propagation of the image fine details to the final layers. 

As shown in the two tasks setting presented in Figure \ref{fig:network-arch}, a multi-task network has one input and {\small $T$} ({\small $T=2$} in the demonstrated example) outputs. All  outputs share the same layers on the contracting path (cf. gray blocks on the left of the network). Multi-task learning blocks are {inserted as last few blocks of} the expansive path to allow learning different features for different tasks. {The bottom blocks on the expansive path, however, remain a single task manner as in the U-net architecture \cite{ronneberger2015u}.}
Each multi-task learning block consists of layers at the same resolution and has two inputs/outputs for the segmentation task and the approximation task respectively (cf. Figure \ref{fig:network-arch}(b)) . 
We note that the amount of labels available for the segmentation tasks is much less than the ones for the approximation task. To prevent over-fitting of the segmentation, the weights of the first layers on the two paths (cf.  blocks in brown) are shared. A residual unit is placed after the brown block for the segmentation task. This gives flexibility for the network to predict segmentation masks. 
The last multi-task block is followed by output layers for the tasks. 

In practice, we use $6$ levels of spatial resolutions, i.e., there are $6$ levels of gray blocks on the contracting path (cf. Figure \ref{fig:network-arch}(a)). 
On the expansive path of the network, we insert {\small $4$} multi-task blocks (cf. blue blocks in Figure \ref{fig:network-arch}(a)).
Moreover, in each gray block, we include a convolutional layer, a batch normalization layer and a leaky ReLU activation function (the slope for the negative part is chosen as {\small $0.01$}). 
Max-pooling layers with factor {\small $2\times 2$} and bilinear interpolation upsampling layers are employed on the contracting path and the expansive path respectively. 

The network architecture in Figure \ref{fig:network-arch} only shows two tasks, but it is possible to extend it if more tasks are included. For instance, if there is a third task for finding certain parts of object boundaries, then one could add layers for this task on the top of the segmentation task feature maps which contain object boundary information.

\section{Experiments}
\begin{figure}[ht]
\centering
\setlength{\tabcolsep}{1pt}
\begin{tabular}{ccc}
\footnotesize{Task 1} & \footnotesize{Task 2} & \footnotesize{Task 3} \\
\hline
\hline
\addlinespace[1ex]
\includegraphics[width=0.25\linewidth]{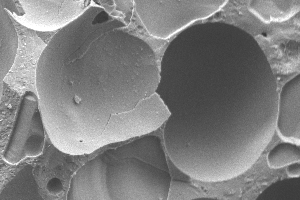} &
\includegraphics[width=0.25\linewidth]{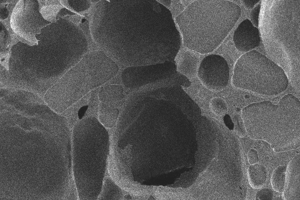} &
\includegraphics[width=0.25\linewidth]{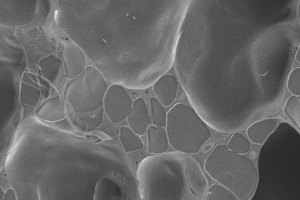} \\
\includegraphics[width=0.25\linewidth]{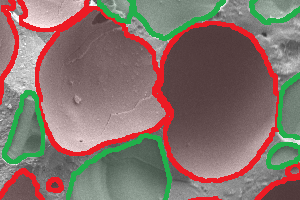} &
\includegraphics[width=0.25\linewidth]{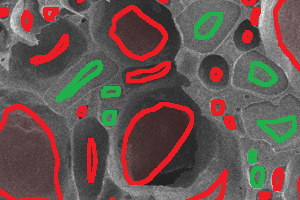} &
\includegraphics[width=0.25\linewidth]{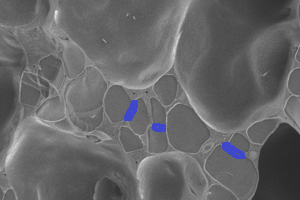} \\
\addlinespace[-1ex]
\multicolumn{3}{c}{\footnotesize (a) SEM images of ice cream}\\
\addlinespace[1ex]
\includegraphics[width=0.25\linewidth]{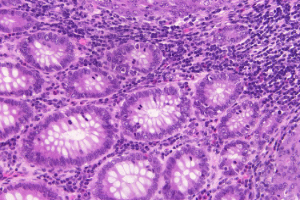} &
\includegraphics[width=0.25\linewidth]{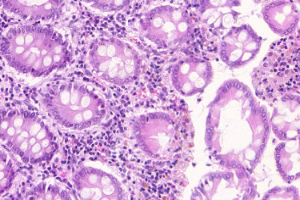} &
\includegraphics[width=0.25\linewidth]{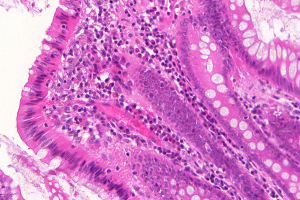} \\
\includegraphics[width=0.25\linewidth]{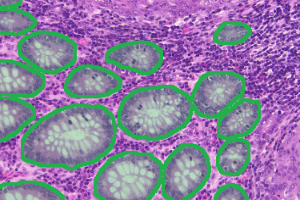} &
\includegraphics[width=0.25\linewidth]{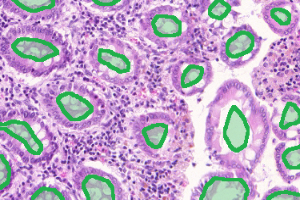} &
\includegraphics[width=0.25\linewidth]{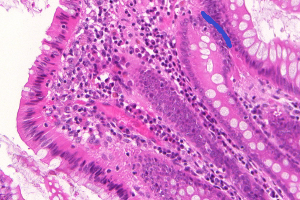} \\
\addlinespace[-1ex]
\multicolumn{3}{c}{\footnotesize (b) H\&E-stained images for gland}\\
\addlinespace[1ex]
\includegraphics[width=0.25\linewidth]{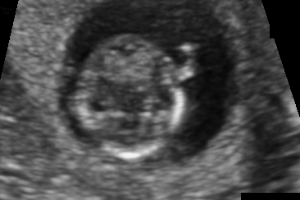} &
\includegraphics[width=0.25\linewidth]{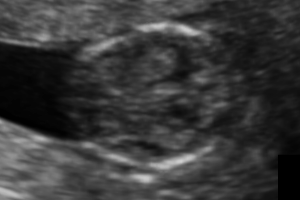} & \\
\includegraphics[width=0.25\linewidth]{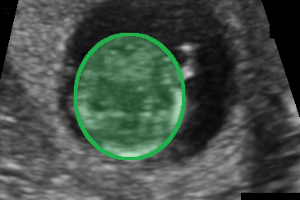} &
\includegraphics[width=0.25\linewidth]{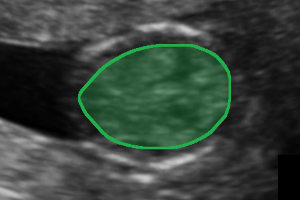} & \\
\addlinespace[-1ex]
\multicolumn{3}{c}{\footnotesize (c) Ultrasound images for fetal head}\\
\end{tabular}
\caption{The annotations demonstrated for three different datasets. In each subfigure, the top row displays the image and the bottom row contains the labels. Each column represents one task.  
Note that the images labeled for different tasks (i.e., images in different columns) can be distributed differently across the whole training set. Typically, only a small portion of the training set is strongly labeled for task 1 (cf. first column), and the majority of the images are weakly labeled (cf. second and the third columns).
In subfigure (a) the red color denotes air bubbles and the green color denotes ice crystals. In subfigures (b) and (c), there is only one class of interest indicated by green regions. 
}\label{fig:annotationdemos}
\end{figure}
The experiments are carried out on three datasets corresponding to distinct applications and involving images acquired in different settings. 
To deal with the associated segmentation problems, we employ our multi-task learning method using both strong labels (SL) and weak labels (WL). 
Examples of labels for the datasets, namely scanning electron microscopy (SEM) images of ice cream, H\&E-stained image dataset for gland and ultrasound images for fetal heads, are presented in Figure \ref{fig:annotationdemos} (a), (b), and (c) respectively. 

We consider different numbers of tasks for the mentioned datasets, depending on the nature of the underlying images. In the fetal head dataset, each image contains one category of objects of interest. Only performing  the \textit{segmentation} task 1 and the \textit{approximation} task 2 is enough.  In the other two datasets, the images may present small-scales and densely distributed objects. An additional \textit{separation}  task 3 is then considered to distinguish connected instances without clear boundaries. 

We also consider different amounts of labels for the tasks. The subset of images labeled for task 1 is always small, accounting for $4.4\%$\textasciitilde $10\%$ of the whole set of images. Task 2 and task 3 require WL that are cheaper to generate, therefore most of the images are labeled for these two tasks. In the rest of this section, unless specified otherwise, we use the ratio of SL and WL summarized in Table \ref{table:SLWL} for our multi-task learning method. 

\begin{table}[h!]
\centering
\caption{The ratio of SL and WL for three datasets}

\begin{tabular}{c|ccc}
\hline
& SL (task 1) & WL (task 2) & WL (task 3) \\
\hline
\hline
SEM & 10\% & 75\% & 100\% \\
H\&E & 9.4\% & 100\% & 100\% \\
Ultrasound & 4.4\% & 100\% & NA\\
\hline
\end{tabular} \label{table:SLWL}
\end{table}

\noindent{\bf  Training.}
Throughout all experiments, we use the network architecture introduced in subsection \ref{subs:network} for our multi-task approach. 
We divide the images into (overlapping) patches for training the network. Augmentation methods such as random flipping, random rotations and random rescaling are performed. 
{For the SEM and ultrasound datasets, we use a crop size of $256 \times 256$ and a batch size of $32$, while for the H\&E gland dataset the size the cropped images and the batch size are set as $480 \times 480$ and $10$ respectively.}
The networks are trained using the Adam optimization method with a learning rate of $0.0002$.  In our multi-task approach, $10$ iterations of the recursive approximation are carried out, and the parameter $\beta$ in the definition of the level set function \eqref{eq:levelset} is set to $1$. 
{For each recursive approximation iteration, $12000$ Adam optimization steps are carried out.}

The final segmentation result is  obtained by performing a last region growing at the approximations provided at the final iterations (i.e., the $10^{\rm th}$ iteration). Specifically, for a given test image, we first obtain the outputs for Task 1 and Task $2$ respectively, and then the masks are computed using the recursive formula defined in equations \eqref{eq:R}, \eqref{eq:levelset} and \eqref{eq:recursive}. A larger value {\small $\beta=100$} is here considered to enforce coherence between predictions of Tasks 1 and 2.

\subsection{SEM images of ice cream}
The segmentation of  different components of food materials is an important step for analyzing their microstructures.  We consider the segmentation of Scanning Electron Microscopy (SEM) images of ice cream. Ice creams, like many other types of foods, are soft solid with complex structures. The segmentation of their images is a challenging problem. Basic components of ice cream are ice crystals, air bubbles and unfrozen matrix. Examples of images are displayed in the top row of Figure \ref{fig:annotationdemos}, for which the annotations are placed in the second row (green for ice crystals, red for air bubbles). 
{Note that in Task 3 we use rough scribbles (cf. the blue color on the third column of Figure \ref{fig:annotationdemos} (a)) to specify the connected object instances without clear boundaries between them. This gives extra data for the DCNN to learn features for these less visible boundaries. We however ignore labeling the clear boundaries shared by touching instances to reduce annotation cost.}

The dataset for this segmentation problem includes $38$ images with large field of views, including a total number of more than $10000$ object instances (air bubbles and ice crystals). The whole dataset is divided into $53\%$ for training, $16\%$ for validation and $31\%$ for testing. 

The ratio of the SL and WL that are used to train our multi-task network is given in Table \ref{table:SLWL}.
The labels being fed to task 2 are rough initial regions obtained manually for approximating the segmentation masks, and they are refined during recursive approximation iterations as introduced in subsection \ref{subs:recursive}.

\begin{figure}[h!]
\centering
\setlength{\tabcolsep}{1pt}
\begin{tabular}{ccccc}
\includegraphics[width=0.15\linewidth]{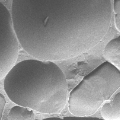} &
\includegraphics[width=0.15\linewidth]{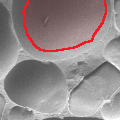} &
\includegraphics[width=0.15\linewidth]{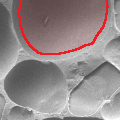} &
\includegraphics[width=0.15\linewidth]{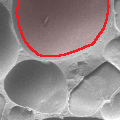} &
\includegraphics[width=0.15\linewidth]{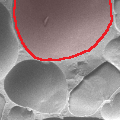}\\
\includegraphics[width=0.15\linewidth]{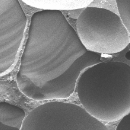} &
\includegraphics[width=0.15\linewidth]{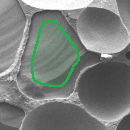} &
\includegraphics[width=0.15\linewidth]{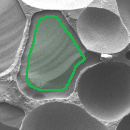} &
\includegraphics[width=0.15\linewidth]{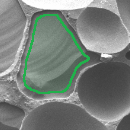} &
\includegraphics[width=0.15\linewidth]{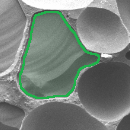} \\
\includegraphics[width=0.15\linewidth]{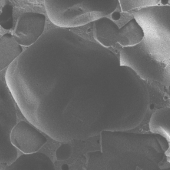} &
\includegraphics[width=0.15\linewidth]{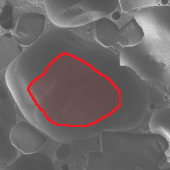} &
\includegraphics[width=0.15\linewidth]{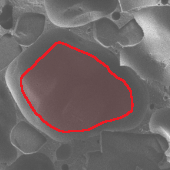} &
\includegraphics[width=0.15\linewidth]{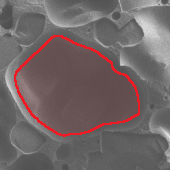} &
\includegraphics[width=0.15\linewidth]{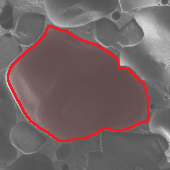} \\
\includegraphics[width=0.15\linewidth]{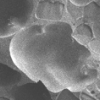} &
\includegraphics[width=0.15\linewidth]{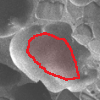} &
\includegraphics[width=0.15\linewidth]{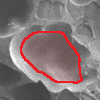} &
\includegraphics[width=0.15\linewidth]{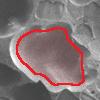} &
\includegraphics[width=0.15\linewidth]{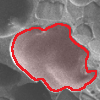}\\
\includegraphics[width=0.15\linewidth]{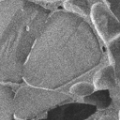} &
\includegraphics[width=0.15\linewidth]{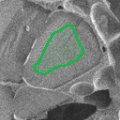} &
\includegraphics[width=0.15\linewidth]{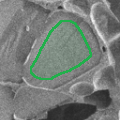} &
\includegraphics[width=0.15\linewidth]{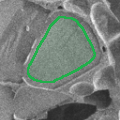} &
\includegraphics[width=0.15\linewidth]{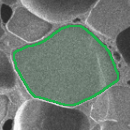} \\
\includegraphics[width=0.15\linewidth]{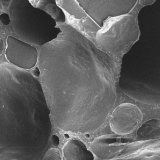} &
\includegraphics[width=0.15\linewidth]{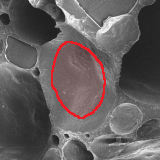} &
\includegraphics[width=0.15\linewidth]{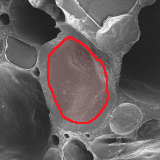} &
\includegraphics[width=0.15\linewidth]{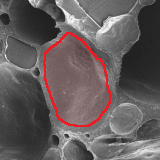} &
\includegraphics[width=0.15\linewidth]{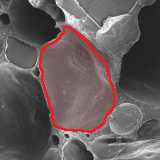} \\
\footnotesize{\textbf{SEM images}} & \footnotesize{\textbf{Iteration 0}} & \footnotesize{\textbf{Iteration 1}} & \footnotesize{\textbf{Iteration 2}} & \footnotesize{\textbf{Iteration 10}}\\
\end{tabular}
\caption{The computed approximations (i.e., from the recursive formula) for the ice cream SEM training images. 
The red color represents the air bubbles while the green one is for ice crystals.}\label{fig:icecream_approximation}
\end{figure}

Results of the computed approximation regions are illustrated in Figure \ref{fig:icecream_approximation}. Though the ground truth segmentation masks are not available for those images, the algorithms achieve much better approximation to the mask at the $10^{\rm th}$ iteration.

We further study the performance of the method by comparing it to several baselines.  The accuracy of the methods will be measured by the dice score 
\begin{equation}\label{eq:dice_definition}
    \frac{ 2 \sum_i g_i p_i }{ \sum_i \qut{g_i + p_i} }
\end{equation}
where $g$ and $p$ are the ground truth mask and the predicted mask respectively. 
The dice scores of the methods are reported  in Table \ref{tab:single-task}. The baseline U-net  is trained on the available small amount of SL. The accuracy of this SL baseline is around 14\% worse than the proposed approach. This implies that  $10\%$ SL is not enough for the fully supervised network to generalize well on the test set. We then consider a single task learning based on only the approximate regions (available for 75\% of the training images). As the labels are very inaccurate (object boundaries are missing from those rough regions), the baseline does not work well either, having an accuracy of $0.837$ for air bubbles and $0.796$ for ice crystals. For the proposed learning method with the recursive approximation task, the dice scores ($0.960$ for air bubbles and $0.953$ for ice crystals) are much better than that of both baselines, thanks to the recursive improvement of the original coarse masks.

\begin{table}[!t]
	\footnotesize
	\caption{Dice scores of segmentation results on the test set of the SEM images. WL baseline is trained on the approximate masks (without a recursive approximation) and the SL baseline is trained on the $10\%$ available ground truth segmentation masks}
	\label{tab:single-task}
	\centering
	\begin{tabular}{c|ccc}
		\toprule
		The models  & air bubbles & ice crystals & Overall \\
		\hline
		WL baseline & 0.725& 0.706 & 0.716 \\
		SL baseline & 0.837 & 0.794 & 0.818 \\
		Multi-task approach &  \textbf{0.960}& \textbf{0.943} & \textbf{0.953} \\
		\bottomrule
	\end{tabular}
\end{table}

\subsection{H\&E-stained image dataset for gland segmentation}
The second dataset we consider in this work is the Glas dataset \cite{sirinukunwattana2017gland}, a benchmark for the segmentation of gland. The glandular structure and morphology can vary remarkably under different histologic grades, and obtaining high quality automatic segmentation is a challenging problem. 
The dataset provides $165$ images acquired from colorectal cancer patients and cover $52$ visual fields for glands in both benign and malignant cases. Ground truth annotations were collected manually by expert pathologists. We follow the common split of this dataset, which includes $85$ images for training, $60$ images for Test Part A and $20$ images for Test Part B. 

Though ground truth annotations are available for the whole training set, we do not use many of them, but instead rely on the rough annotations for the majority of images as shown in the last two columns of Figure \ref{fig:annotationdemos} (c).
Our multi-task algorithm is run with those labels and then compared with the state of the art methods on this dataset, that are  {\em trained using all available ground truth annotations}{, as well as other weakly supervised methods.}

\begin{figure}[h!]
\centering
\includegraphics[width=0.8\linewidth, trim=10 5 20 30 ,clip]{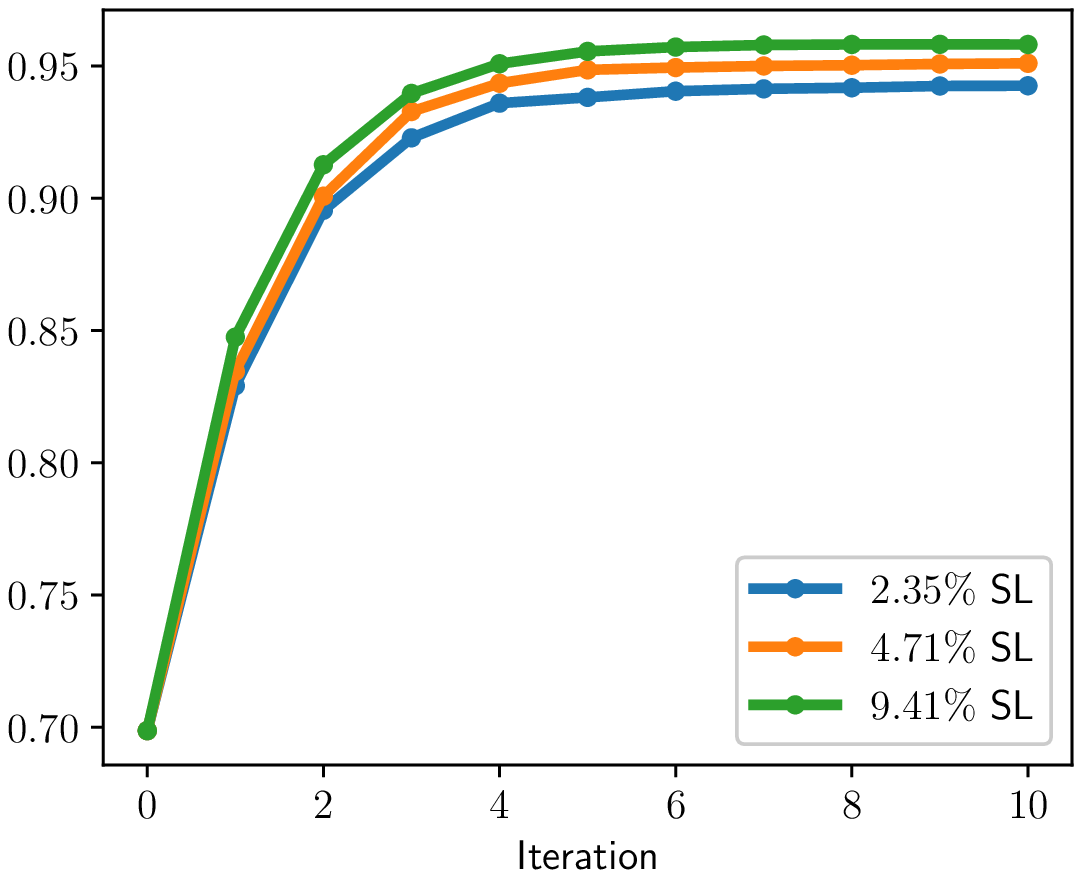}
\vspace{-0.4cm}
\caption{Evolution to the dice scores with respect to the approximation iterations for the first $10$ iterations on the  H\&E-stained image dataset}\label{fig:glanddicevsiteration}
\end{figure}
The comparisons are done with different amounts of strongly labeled images. We start from $2$ images ($2.4\%$ of the training set) and increase it up to $8$ images ($9.4\%$). We run $10$ iterations of the recursive approximations. After each iteration; approximate segmentation masks are updated for the whole training set.  The dice scores for those approximations averaged over the $85$ training images are plotted against the iteration number in Figure \ref{fig:glanddicevsiteration}. In case of $2$ images with strong labels, the mean dice score starts from around $0.7$ and is increased over the iterations to around $0.94$ at the $10^{\rm th}$ iteration. When  $8$ images with strong labels are considered, the score is slightly better, reaching $0.96$ at the last iteration.  

\begin{figure}[h!]
\centering
\setlength{\tabcolsep}{1pt}
\begin{tabular}{cccc}
\sidecap{\footnotesize Iteration 0}
&
\includegraphics[width=0.30\linewidth, trim=16 67 83 0, clip]{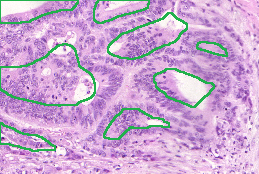} &
\includegraphics[width=0.30\linewidth, trim=50 67 50 0, clip]{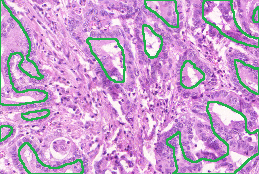} &
\includegraphics[width=0.30\linewidth, trim=0 0 100 67, clip]{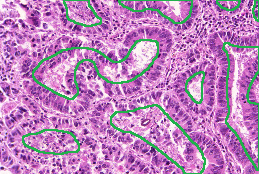} \\
\sidecap{\footnotesize Iteration 1}
&
\includegraphics[width=0.30\linewidth, trim=16 67 83 0, clip]{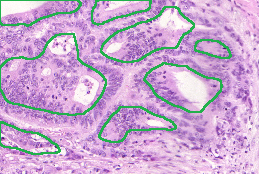} &
\includegraphics[width=0.30\linewidth, trim=50 67 50 0, clip]{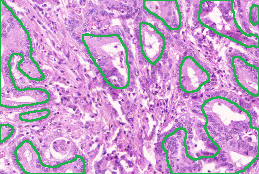} &
\includegraphics[width=0.30\linewidth, trim=0 0 100 67, clip]{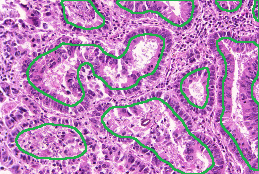} \\
\sidecap{\footnotesize Iteration 5}
&
\includegraphics[width=0.30\linewidth, trim=16 67 83 0, clip]{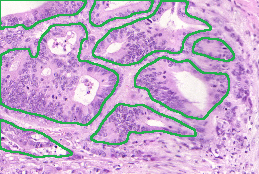} &
\includegraphics[width=0.30\linewidth, trim=50 67 50 0, clip]{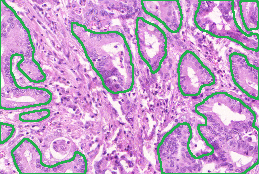} &
\includegraphics[width=0.30\linewidth, trim=0 0 100 67, clip]{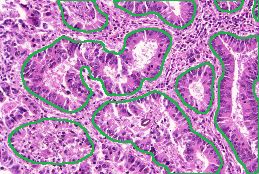} \\
\sidecap{\footnotesize Iteration 10}
&
\includegraphics[width=0.30\linewidth, trim=16 67 83 0, clip]{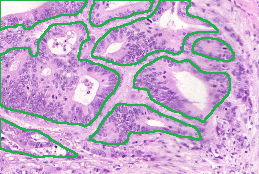} &
\includegraphics[width=0.30\linewidth, trim=50 67 50 0, clip]{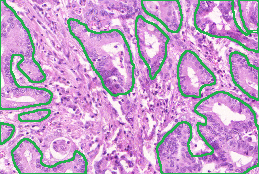} &
\includegraphics[width=0.30\linewidth, trim=0 0 100 67, clip]{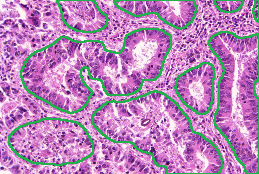} \\
\sidecap{\footnotesize Ground truth}
&
\includegraphics[width=0.30\linewidth, trim=16 67 83 0, clip]{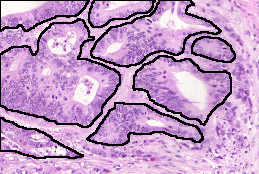} &
\includegraphics[width=0.30\linewidth, trim=50 67 50 0, clip]{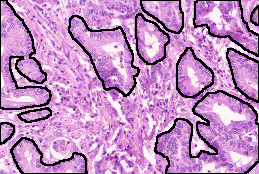} &
\includegraphics[width=0.30\linewidth, trim=0 0 100 67, clip]{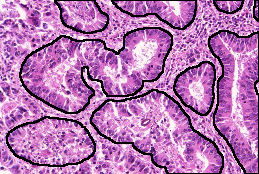} 
\end{tabular}
\caption{The evolution of the contours (best viewed in color). The \textbf{green} curves in the top $4$ rows are the contours corresponding to iteration $0$, $1$, $5$ and $10$ respectively, and the curve in \textbf{black} at the bottom row is ground truth contour.}\label{fig:glandevolve}
\end{figure}

Some qualitative results are given in Figure \ref{fig:glandevolve}. The evolution of the contour lines through training iterations is illustrated and compared with the ground truth contours. We underline that ground truth segmentation masks of these images have not been used during the training and the updates of the contour are determined by the recursive approximation task only. This demonstrates that starting from rough initial regions, the algorithm is able to estimate correctly complex object  boundaries in  textured images. 

The segmentation results are reported in Table \ref{tab:score_gland}. The quality of the segmentation is measured by two metrics, namely the object dice index and the Hausdorff distance. {Note that in comparison to the dice score defined in \eqref{eq:dice_definition}, the object dice index reflects how well the predicted region of each individual object instances matches the one in ground truth annotations rather than just a class-wise match.} The object dice index takes values between $0$ and $1$ where $1$ means a perfect match. The Hausdorff distance measures the distance between the predicted object boundaries and the ground truth boundaries. We refer the readers to \cite{sirinukunwattana2017gland} for the definitions of the two metrics. As shown in the table, with only 2 images ($2.4$\%) strongly labeled, the method already gives reasonable segmentation results, with  object dice indexes of $0.842$ and $0.739$ for test Part A and test Part B respectively. 

{
Pseudo label based approaches (see e.g., \cite{khoreva2017simple, jing2019coarse}), that use fake masks for supervision, is another class of weakly supervised learning methods. 
We compare our method with the learning method based on pseudo labels. Following the work of \cite{khoreva2017simple}, the pseudo labels are obtained by the Grabcut method \cite{rother2004grabcut}. In our implementation, for each initial region we assign a sufficiently large bounding box as the region of interest. By setting the initial region as foreground and the pixels far away from the initial regions as background, Grabcut is applied to get the mask of the object. The pseudo segmentation mask is created by repeating the algorithm over all the objects. For strongly labeled images, the ground truth segmentation masks are used instead of the pseudo masks. 
The segmentation accuracy of PL is nevertheless significantly lower than the proposed method given the same amount of annotations, as shown in Table \ref{tab:score_gland}.}

{We also report results of fully supervised methods, including U-net (Freiburg \cite{ronneberger2015u}), the multi-path convolutional neural network (ExB \cite{sirinukunwattana2017gland}), the deep contour-aware network (CUMedVision \cite{chen2016dcan}), a deep multichannel neural network exploiting regional, location and boundary cues (Xu et al. \cite{xu2017gland}), and a modified LinkNet incorporating invariant local binary pattern (Rezaei et al \cite{rezaei2019gland}).} 
Using $9.4$\% SL, our approach reaches an accuracy comparable to some of the best fully supervised methods on this dataset, that consider $100\%$ SL.

\begin{table}[ht!]
\centering
\caption{Comparison with two metrics for the Glas dataset. Results on both test subsets are reported respectively for the weak supervision (\textbf{WS}) methods and strong supervision (\textbf{SS}) methods. The WS methods use WL as well as SL (available for 2.4\% \textasciitilde 9.4\% images). All the SS methods use 100\% SL.}\label{tab:score_gland}
\begin{tabular}{|c|c|cc|cc|}
\hline
\multirow{2}{*}{} & \multirow{2}{*}{Methods} & \multicolumn{2}{c|}{Object dice index} & \multicolumn{2}{c|}{Hausdorff distance } \\
\cline{3-6}
& & Part A & Part B & Part A & Part B \\
\hline
\hline
\multirow{6}{*}{\rotatebox[origin=c]{90}{\textbf{WS}}}
& Ours (2.4\% SL+WL) &  0.843 & 0.739 &  63.90 & 142.65  \\
& PL (2.4\% SL+WL) & 0.716 & 0.646 & 130.26 & 247.26 \\
\cline{2-6}
& Ours (4.7\% SL+WL)  &  0.853 & 0.815 & 61.11 & 115.39  \\
& PL (4.7\% SL+WL) & 0.732 & 0.628 & 127.50 & 259.81 \\
\cline{2-6}
& Ours (9.4\% SL+WL) &  \textbf{0.881} & \textbf{0.828} & 
\textbf{50.53} & \textbf{112.92}\\
& PL (9.4\% SL+WL) & 0.724 & 0.630  & 137.16 & 275.93 \\
\hline
\hline
\multirow{5}{*}{\rotatebox[origin=c]{90}{\textbf{SS}}}
& Freiburg \cite{ronneberger2015u} & 0.876 & 0.786 & 57.10 & 148.46\\
& ExB \cite{sirinukunwattana2017gland} &  0.882 & 0.786 &  57.41  & 145.58\\
& CUMedVision \cite{chen2016dcan} & 0.897 & 0.781 & 45.42 & 160.35\\
& Xu et al. \cite{xu2017gland} & \textbf{0.908} & \textbf{0.833} & \textbf{44.13} & 116.82\\
& Rezaei et al. \cite{rezaei2019gland} & 0.867 & 0.822 & 44.74 & \textbf{96.98} \\
\hline
\end{tabular}
\end{table}

 \subsection{Ultrasound images for fetal head}
Ultrasound imaging is one of the most commonly used techniques for monitoring and estimating the growth and health of  fetuses during pregnancy.  In this experiment we study the problem of segmenting the fetal head from ultrasound images. The images are from a database of Radboud University Medical Center \cite{van2018automated}. We take a set of $999$ images from the database, which we further split into $58\%$ images for training, $17\%$ images for validation, and $25\%$ for testing. The ground truth annotations for this dataset were provided by experienced sonographers. 
 
As the regions of interest (i.e., fetal heads) are well isolated in the fetal head ultrasound images,  we do not perform the third separating task that was designed for the first experiments. We used for the training a small portion ($4.4\%$ i.e., 25 images) of the ground truth annotation set as well as the weak labels shown in the second column of Figure \ref{fig:annotationdemos}(c), which do not describe the accurate boundaries. Since there are no known existing weakly supervised deep learning approaches developed on this dataset, we compared our method with a fully supervised baseline (U-net trained on $100\%$ SL). 

\begin{figure}[h!]
	 \centering
		 \begin{tabular}{c}
		 \includegraphics[width=0.6\textwidth, trim = 20 15 35 20, clip]{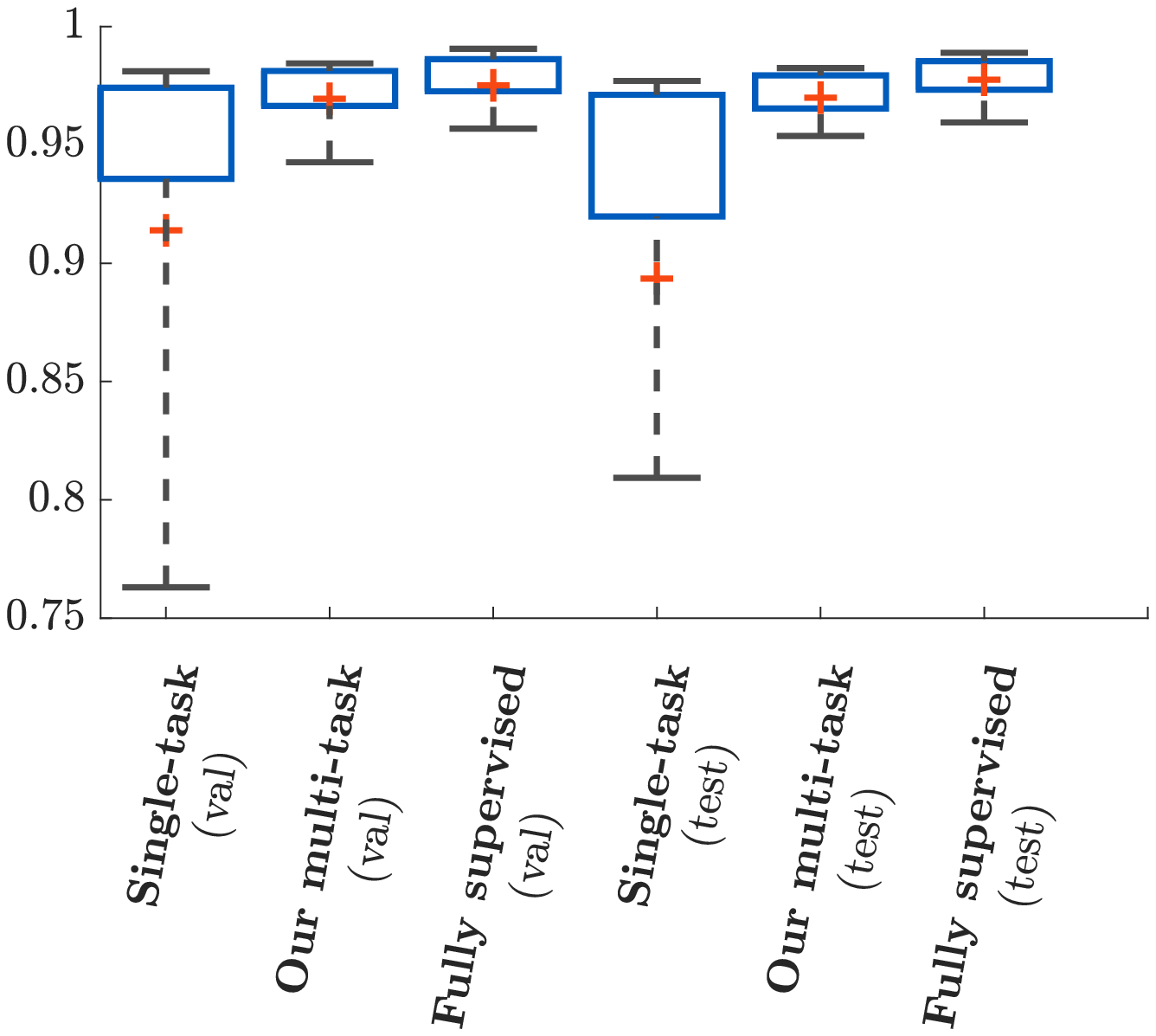}
		 \end{tabular}
		 \vspace{-0.2cm}
		 \caption{Dice score of the three compared methods on ultrasound images split into a validation set and a test set. The baseline single task method learns from $4.4\%$ SL. Our multi-task approach uses the $4.4\%$ SL for the segmentation task while the approximation task sees only WL. The fully supervised scheme is however trained on a full set of SL. The red plus symbol denotes the mean score and the box covers the middle 80\% of the score distribution.} \label{fig:fh_dice}
 \end{figure}

We first train a single task network using only the small portion (4.4\%) of the images with ground truth annotations. The accuracy of this model measured in the dice score is plotted in Figure \ref{fig:fh_dice} for both the validation and test sets. The low scores for part of the images show that the amount of SL is not sufficient for the baseline model to generalize well. The mean dice scores are $0.912$ and $0.908$ on the validation set and test set respectively.  Our multi-task approach, in contrast, results in scores of $0.970$ (validation) and $0.970$ (test). We underline that less than $1\%$ is gained with the supervised training on the full set of SL, that reaches dice scores of $0.975$ and $0.977$ respectively. 

\begin{figure}[h!]
\centering
\setlength{\tabcolsep}{1pt}
\begin{tabular}{cccc}
\sidecapcustom{2cm}{\footnotesize Image}
&
\includegraphics[width=0.25\linewidth, trim=0 0 0 0, clip]{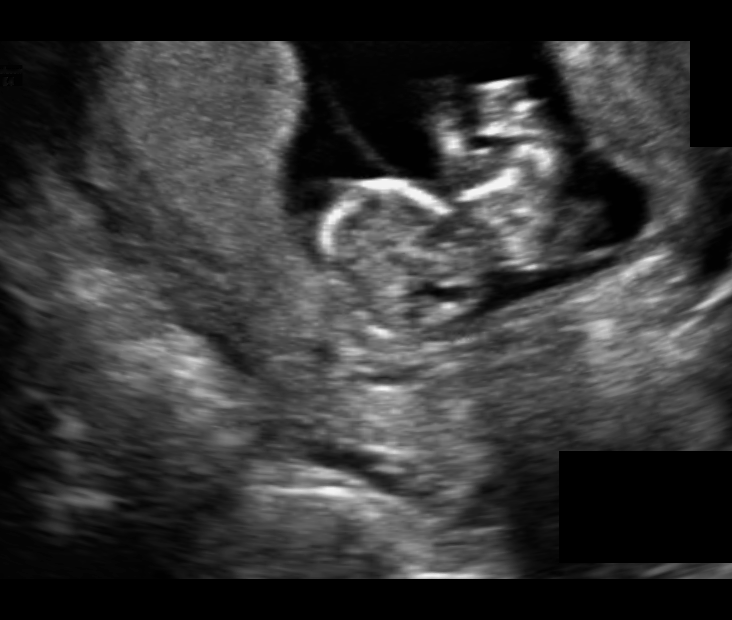} &
\includegraphics[width=0.25\linewidth, trim=0 0 0 0, clip]{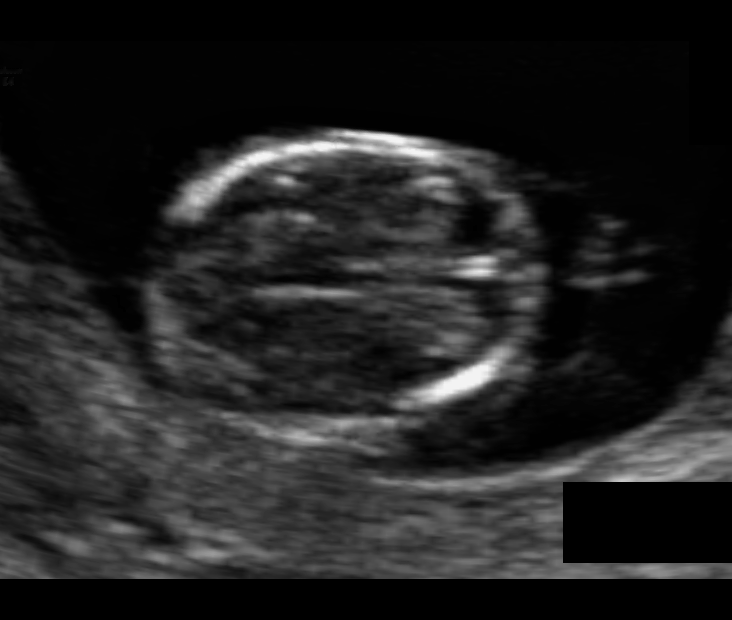} &
\includegraphics[width=0.25\linewidth, trim=0 0 0 0, clip]{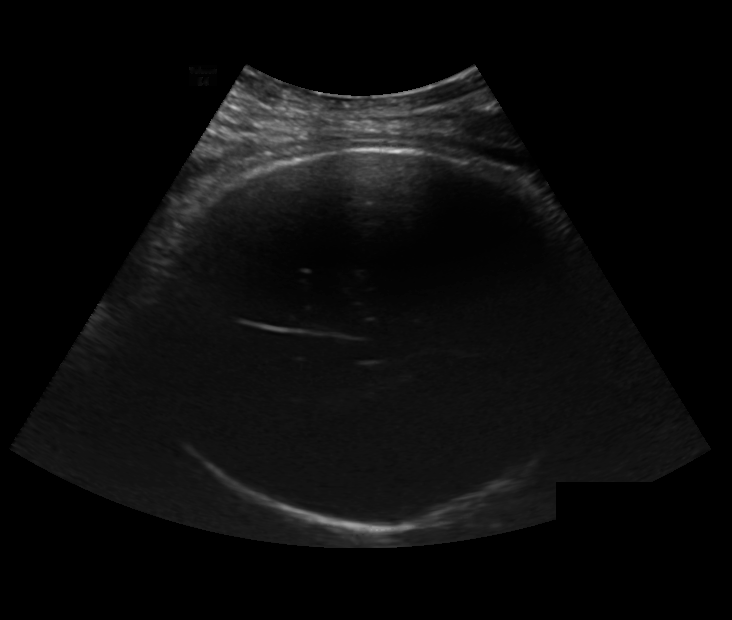} 
\\
\sidecapcustom{2cm}{\footnotesize Baseline}
&
\includegraphics[width=0.25\linewidth, trim=0 0 0 0, clip]{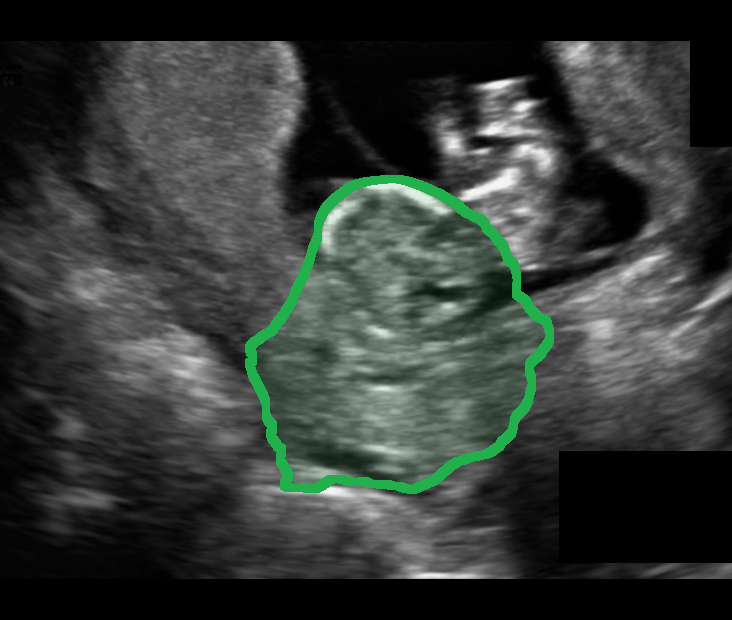} &
\includegraphics[width=0.25\linewidth, trim=0 0 0 0, clip]{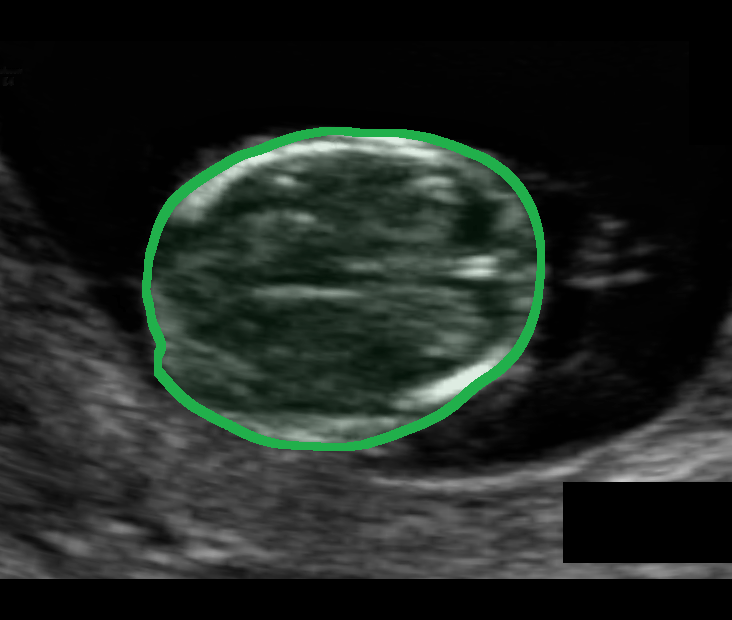} &
\includegraphics[width=0.25\linewidth, trim=0 0 0 0, clip]{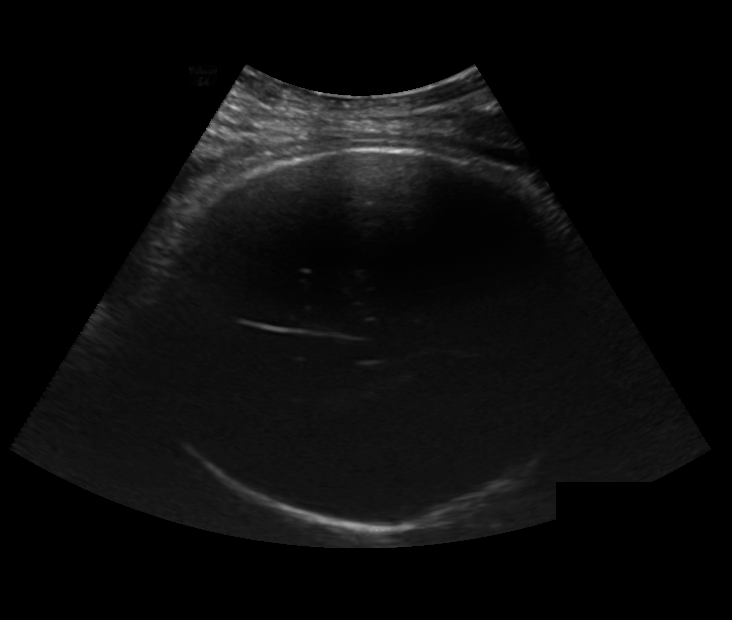} 
\\
\sidecapcustom{2cm}{\footnotesize Multi-task}
&
\includegraphics[width=0.25\linewidth, trim=0 0 0 0, clip]{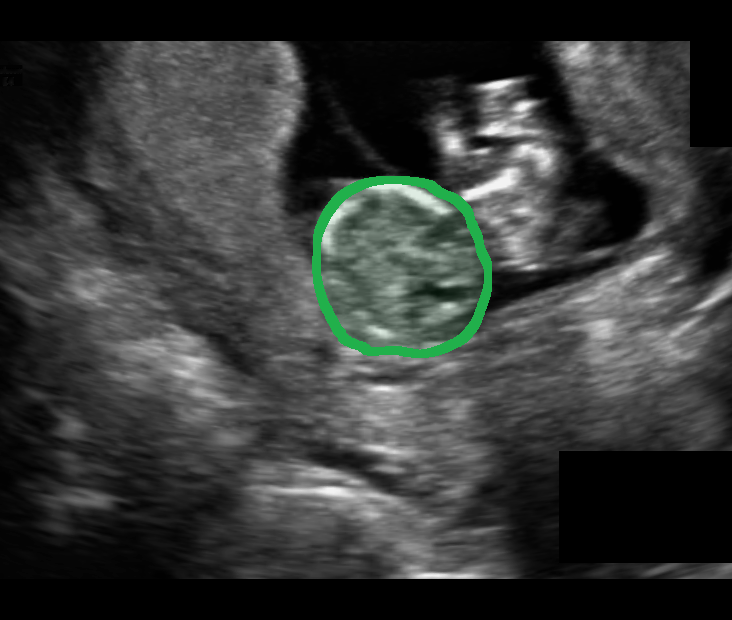} &
\includegraphics[width=0.25\linewidth, trim=0 0 0 0, clip]{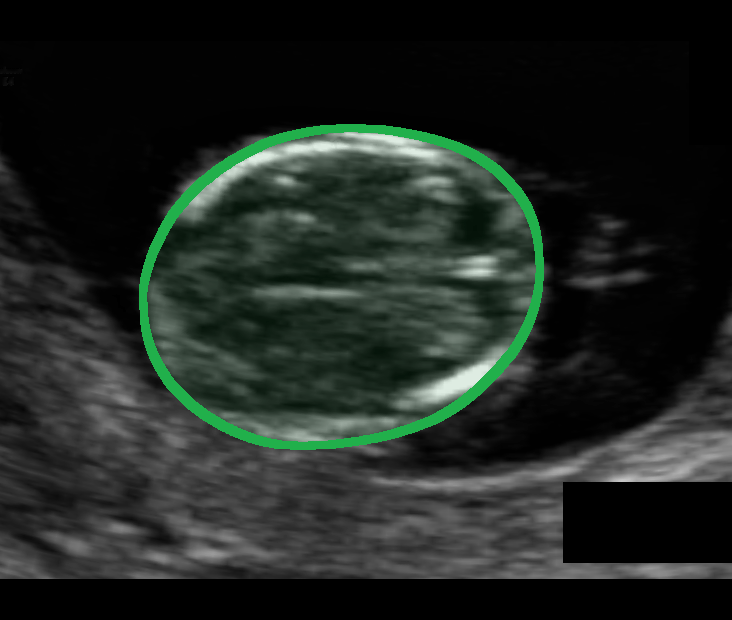} &
\includegraphics[width=0.25\linewidth, trim=0 0 0 0, clip]{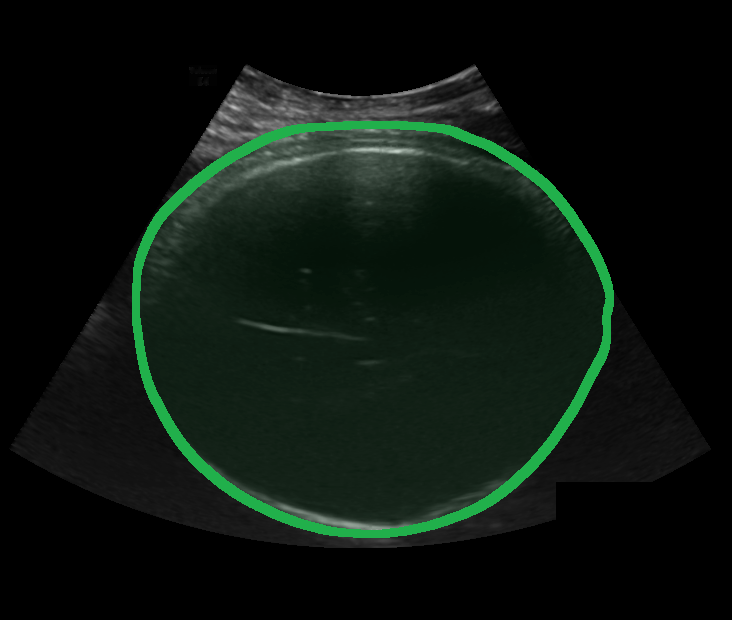} 
\\
\sidecapcustom{2cm}{\footnotesize ~Ground truth}
&
\includegraphics[width=0.25\linewidth, trim=0 0 0 0, clip]{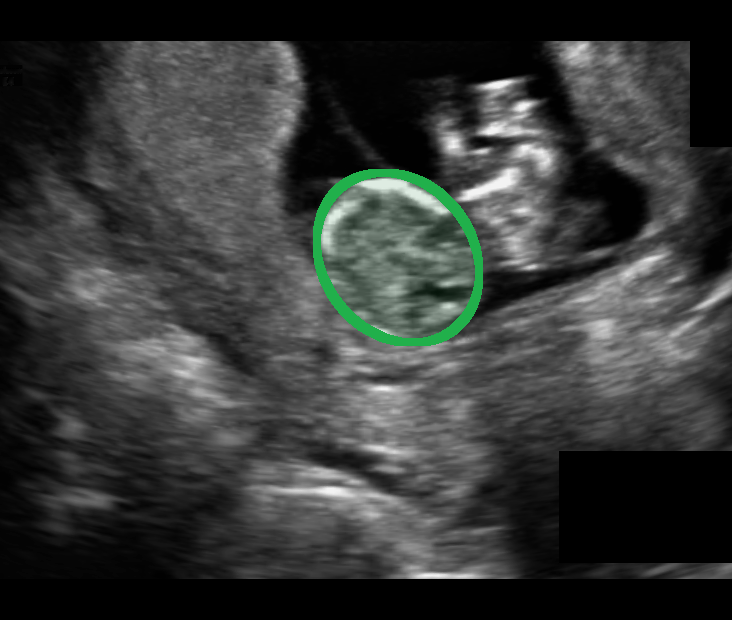} &
\includegraphics[width=0.25\linewidth, trim=0 0 0 0, clip]{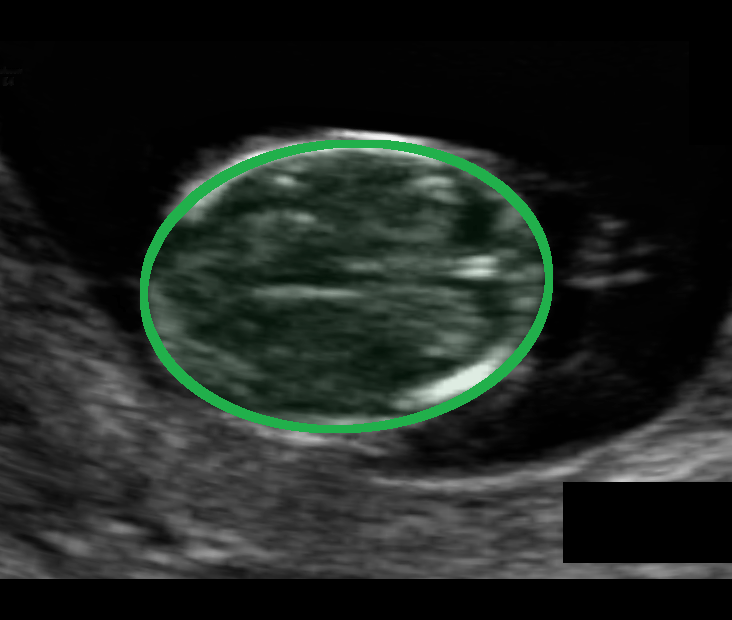} &
\includegraphics[width=0.25\linewidth, trim=0 0 0 0, clip]{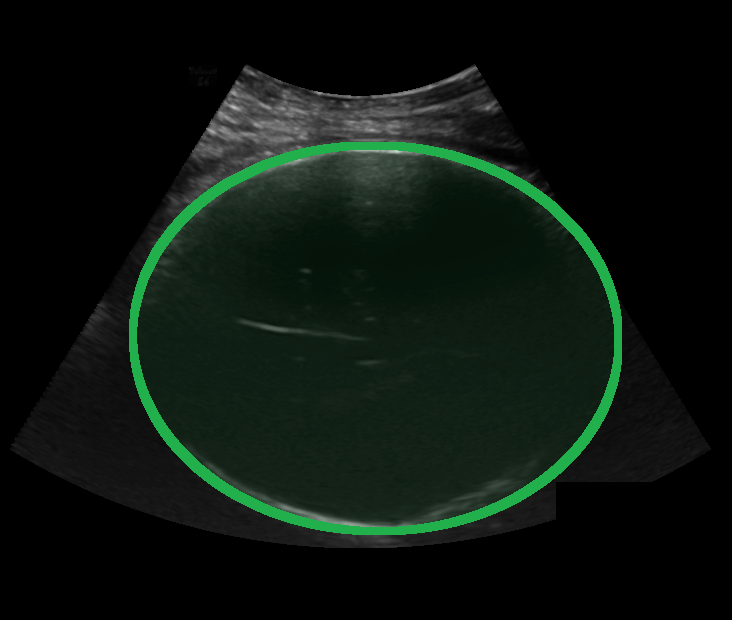} 
\\
\end{tabular}
\caption{The segmentation result of the baseline method (second row) and our approach (third row) demonstrated together with the input image (first row) and the ground truth annotations (last row).  The baseline method is fed with $4.4\%$ SL, and our approach uses $4.4\%$ SL plus WL.}\label{fig:hc_demo}
\end{figure}

In Figure \ref{fig:hc_demo}, examples of segmentation results are illustrated. Due to the lack of SL for training, the results of the baseline method can  be unreliable: over-estimation of the regions or completely left out of the fetal head. The proposed method fits the contours much better, even though contours are not directly provided in WL. 

\if1
\begin{table}[h]
	\footnotesize
	\caption{multi-task learning results. The dice index for the two classes are displaced. The item for which the corresponding object classes account for less than 5\% is neglected and replaced by $/$.}
	\label{tab:multi-task}
	\centering
	\begin{tabular}{c|ccccccccccccc}
		\toprule
		image no. & 1 & 2 & 3 & 4 & 5 & 6 & 7 & 8 & 9 & 10 & 11 & 12 & overall \\
		\hline
		{air bubbles}  & 0.93& 0.97 & 0.87 & 0.97 & 0.95 & 0.93 & 0.92 & 0.92 & $/$ & $/$ & 0.96 & $/$ & 0.93 \\
		{ice crystals} &  0.91& 0.95 & 0.91 & 0.93 & 0.89 & 0.90 & 0.90 & 0.89 & 0.88 & 0.92 & 0.93 & 0.94 & 0.91 \\
		\bottomrule
	\end{tabular}
\end{table}
\fi

\FloatBarrier
\section{Conclusion}
In this work we propose a learning framework for image segmentation that integrates the segmentation problem within a recursive approximation task. This task aims at finding the contours starting from very rough initial regions within a training process. It consists of a set of iterations where approximate regions grow towards object boundaries. The approximation task is performed together with a segmentation task trained on a limited amount of strong labels (ground truth accurate segmentation masks). The approach uses no handcrafted features for the images or prepossessing of the weak labels, and the growing of the regions for the approximation task can be implemented easily.  

The approach is applied to three different datasets successfully. We have demonstrated that it does not consume a large number of strong labels that are usually expensive to collect. 
The method rather relies on $2\%-10\%$ of strong labels together with coarse and cheap annotations for the majority of the images, while reaching an accuracy close to  models trained with accurate strong labels on the full training set.

\bibliographystyle{unsrt}
\bibliography{DL_segmentation_ref.bib}

\end{document}